%% file: 0_main.tex
\documentclass[nohyperref]{article}

\usepackage{microtype}
\usepackage{graphicx}
\usepackage{booktabs} 
\usepackage{hyperref}

\usepackage[accepted]{icml2022}
\usepackage{amsmath}
\usepackage{amssymb}
\usepackage{mathtools}
\usepackage{amsthm}
\usepackage[utf8]{inputenc} 
\usepackage[T1]{fontenc}    
\usepackage{url}            
\usepackage{amsfonts}       
\usepackage{nicefrac}       
\usepackage{xcolor}         
\usepackage{colortbl}
\usepackage{subcaption}
\usepackage{enumitem}
\usepackage{multirow}
\usepackage{capt-of}
\usepackage{wrapfig}
\usepackage{makecell}
\usepackage{bm}
\usepackage{arydshln}
\usepackage{placeins}
\hypersetup{
    colorlinks = true,
    citecolor = blue,
    linkcolor = red
}
\usepackage{comment}
\usepackage{algorithm}
\usepackage{algorithmic}
\usepackage{eqparbox}

\definecolor{gg}{gray}{0.92}
\newcolumntype{a}{>{\columncolor{gg}}c}

\input{math_commands}
\usepackage[capitalize,noabbrev]{cleveref}

\theoremstyle{plain}

\theoremstyle{definition}

\theoremstyle{remark}

\usepackage[textsize=tiny]{todonotes}

\usepackage[hang,flushmargin]{footmisc} 

\icmltitlerunning{Score-based Generative Modeling of Graphs via the System of SDEs}

\begin{document}

\twocolumn[
\icmltitle{Score-based Generative Modeling of Graphs via \\ the System of Stochastic Differential Equations}

\icmlsetsymbol{equal}{*}

\begin{icmlauthorlist}
\icmlauthor{Jaehyeong Jo}{KAIST,equal}
\icmlauthor{Seul Lee}{KAIST,equal}
\icmlauthor{Sung Ju Hwang}{KAIST,AITRICS}
\end{icmlauthorlist}

\icmlaffiliation{KAIST}{Korea Advanced Institute of Science and Technology (KAIST), Seoul, South Korea}
\icmlaffiliation{AITRICS}{AITRICS, South Korea}

\icmlcorrespondingauthor{Jaehyeong Jo}{harryjo97@kaist.ac.kr}
\icmlcorrespondingauthor{Seul Lee}{seul.lee@kaist.ac.kr}
\icmlcorrespondingauthor{Sung Ju Hwang}{sjhwang82@kaist.ac.kr}

\icmlkeywords{Graph Generation, Score-based Generative Modeling}

\vskip 0.3in
]
\printAffiliationsAndNotice{\;\;\icmlEqualContribution} 

\input{1_abstract}
\input{2_introduction}
\input{3_related_work}
\input{4_method}

\input{5_experiments}

\input{6_conclusion}
\bibliography{references}
\bibliographystyle{icml2022}
\input{7_appendix}

\end{document}

%% file: math_commands.tex
\usepackage{amsmath, amsfonts, bm}

\newcommand{\score}[1]{\nabla_{\!\!#1}\!\log p_t(#1)}
\newcommand{\pscore}[2]{\nabla_{\!\!#1}\!\log p_t(#2)}
\newcommand{\dscore}[2]{\nabla_{\!\!\bm{#1}_t}\!\log p_{0t}(\bm{#2}_t|\bm{#2}_0)}
\newcommand{\overbar}[1]{\mkern 1.5mu\overline{\mkern-1.5mu#1\mkern-1.5mu}\mkern 1.5mu}
\newcommand{\fop}{\hat{\mathcal{L}}^{*}}

\renewcommand{\algorithmiccomment}[1]{\hfill{ $\rhd$ #1}}

%% file: 1_abstract.tex
\begin{abstract}
Generating graph-structured data requires learning the underlying distribution of graphs. Yet, this is a challenging problem, and the previous graph generative methods either fail to capture the permutation-invariance property of graphs or cannot sufficiently model the complex dependency between nodes and edges, which is crucial for generating real-world graphs such as molecules. To overcome such limitations, we propose a novel score-based generative model for graphs with a continuous-time framework. Specifically, we propose a new graph diffusion process that models the joint distribution of the nodes and edges through a system of stochastic differential equations (SDEs). Then, we derive novel score matching objectives tailored for the proposed diffusion process to estimate the gradient of the joint log-density with respect to each component, and introduce a new solver for the system of SDEs to efficiently sample from the reverse diffusion process. We validate our graph generation method on diverse datasets, on which it either achieves significantly superior or competitive performance to the baselines. Further analysis shows that our method is able to generate molecules that lie close to the training distribution yet do not violate the chemical valency rule, demonstrating the effectiveness of the system of SDEs in modeling the node-edge relationships. Our code is available at https://github.com/harryjo97/GDSS.
\end{abstract}

%% file: 2_introduction.tex
\section{Introduction}
Learning the underlying distribution of graph-structured data is an important yet challenging problem that has wide applications, such as understanding the social networks~\cite{graphite, graphgan}, drug design~\cite{drug-design/1, drug-design/2}, neural architecture search (NAS)~\cite{architecture-search, metad2a}, and even program synthesis~\cite{program-synthesis}. Recently, deep generative models have shown success in graph generation by modeling complicated structural properties of graphs, exploiting the expressivity of neural networks. Among them, autoregressive models~\cite{you2018graphrnn,liao2019efficient} construct a graph via sequential decisions, while one-shot generative models~\cite{de2018molgan, GNF} generate components of a graph at once. Although these models have achieved a certain degree of success, they also possess clear limitations. Autoregressive models are computationally costly and cannot capture the permutation-invariant nature of graphs, while one-shot generative models based on the likelihood fail to model structural information due to the restriction on the architectures to ensure tractable likelihood computation.

Apart from the likelihood-based methods, \citet{score-based/graph/1} introduced a score-based generative model for graphs, namely, edge-wise dense prediction graph neural network (EDP-GNN). However, since EDP-GNN utilizes the discrete-step perturbation of heuristically chosen noise scales to estimate the score function, both its flexibility and its efficiency are limited. Moreover, EDP-GNN only generates adjacency matrices of graphs, thus is unable to fully capture the node-edge dependency which is crucial for the generation of real-world graphs such as molecules.

To overcome the limitations of previous graph generative models, we propose a novel score-based graph generation framework on a continuous-time domain that can generate both the node features and the adjacency matrix. Specifically, we propose a novel \emph{Graph Diffusion via the System of Stochastic differential equations} (GDSS), which describes the perturbation of both node features and adjacency through a system of SDEs, and show that the previous work of \citet{score-based/graph/1} is a special instance of GDSS. Diffusion via the system of SDEs can be interpreted as the decomposition of the full diffusion into simpler diffusion processes of respective components while modeling the dependency. Further, we derive novel training objectives for the proposed diffusion, which enable us to estimate the gradient of the joint log-density with respect to each component, and introduce a new integrator for solving the proposed system of SDEs.

We experimentally validate our method on generic graph generation tasks by evaluating the generation quality on synthetic and real-world graphs, on which ours outperforms existing one-shot generative models while achieving competitive performance to autoregressive models. We further validate our method on molecule generation tasks, where ours outperforms the state-of-the-art baselines including the autoregressive methods, 
demonstrating that the proposed diffusion process through the system of SDEs is able to capture the complex dependency between nodes and edges. We summarize our main contributions as follows:
\vspace{-0.1in}
\begin{itemize}[itemsep=0.5mm, parsep=1pt, leftmargin=*]
    \item We propose a novel score-based generative model for graphs that overcomes the limitation of previous generative methods, by introducing a diffusion process for graphs that can generate node features and adjacency matrices simultaneously via the system of SDEs.
    \item We derive novel training objectives to estimate the gradient of the joint log-density for the proposed diffusion process and further introduce an efficient integrator to solve the proposed system of SDEs.
    \item We validate our method on both synthetic and real-world graph generation tasks, on which ours outperforms existing graph generative models.
\end{itemize}

%% file: 3_related_work.tex
\section{Related Work}
\paragraph{Score-based Generative Models}
Score-based generative models generate samples from noise by first perturbing the data with gradually increasing noise, then learning to reverse the perturbation via estimating the score function, which is the gradient of the log-density function with respect to the data. Two representative classes of score-based generative models have been proposed by \citet{score-based/0} and \citet{score-based/ddpm}, respectively. Score matching with Langevin dynamics (SMLD)~\cite{score-based/0} estimates the score function at multiple noise scales, then generates samples using annealed Langevin dynamics to slowly decrease the scales. On the other hand, denoising diffusion probabilistic modeling (DDPM)~\cite{score-based/ddpm} backtracks each step of the noise perturbation by considering the diffusion process as a parameterized Markov chain and learning the transition of the chain. Recently, \citet{score-based/2} showed that these approaches can be unified into a single framework, describing the noise perturbation as the forward diffusion process modeled by the stochastic differential equation (SDE). Although score-based generative models have shown successful results for the generation of images~\cite{score-based/ddpm, score-based/2, score-based/3, score-based/4}, audio~\cite{score-based/audio/1, score-based/audio/2, score-based/audio/3, score-based/audio/4}, and point clouds~\cite{score-based/point-clouds/1, score-based/point-clouds/2}, the graph generation task remains to be underexplored due to the discreteness of the data structure and the complex dependency between nodes and edges. We are the first to propose a diffusion process for graphs and further model the dependency through a system of SDEs. It is notable that recent developments of score-based generative methods, such as latent score-based generative model (LSGM)~\cite{lsgm} and critically-damped Langevin diffusion (CLD)~\cite{CLD}, are complementary to our method as we can apply these methods to improve each component-wise diffusion process.

\vspace{-0.1in}
\paragraph{Graph Generative Models}
The common goal of graph generative models is to learn the underlying distribution of graphs. Graph generative models can be classified into two categories based on the type of the generation process: autoregressive and one-shot. Autoregressive graph generative models include generative adversarial network (GAN) models~\cite{you2018graph}, recurrent neural network (RNN) models~\cite{you2018graphrnn, popova2019molecularrnn}, variational autoencoder (VAE) models~\cite{jin2018junction, jin2020hierarchical}, and normalizing flow  models~\cite{shi2020graphaf, luo2021graphdf}. Works that specifically focus on the scalability of the generation comprise another branch of autoregressive models~\cite{liao2019efficient, dai2020scalable}. Although autoregressive models show state-of-the-art performance, they are computationally expensive and cannot model the permutation-invariant nature of the true data distribution. On the other hand, one-shot graph generative models aim to directly model the distribution of all the components of a graph as a whole, thereby considering the structural dependency as well as the permutation-invariance. One-shot graph generative models can be categorized into GAN models~\cite{de2018molgan}, VAE models~\cite{ma2018constrained}, and normalizing flow models~\cite{madhawa2019graphnvp, zang2020moflow}. There are also recent approaches that utilize energy-based models (EBMs) and score-based models, respectively~\cite{liu2021graphebm, score-based/graph/1}. Existing one-shot generative models perform poorly due to the restricted model architectures for building a normalized probability, which is insufficient to learn the complex dependency of nodes and edges. To overcome these limitations, we introduce a novel permutation-invariant one-shot generative method based on the score-based model that imposes fewer constraints on the model architecture, compared to previous one-shot methods.

\begin{figure*}[t!]
    \centering
    \includegraphics[width=0.98\linewidth]{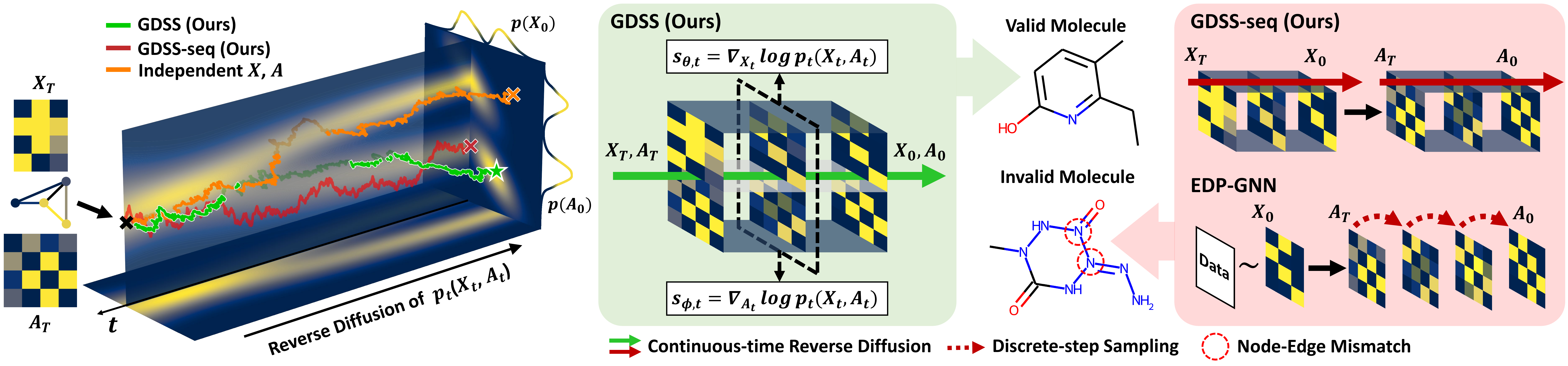}
    \vspace{-0.1in}
    \caption{\small\textbf{(Left) Visualization of graph generation through the reverse-time diffusion process.} The colored trajectories denote different types of diffusion processes in the joint probability space of node features $\bm{X}$ and adjacency $\bm{A}$. We compare three types of diffusion: GDSS (green) can successfully generate samples from the data distribution by modeling the dependency between the components, whereas GDSS-seq (red) or the independent diffusion of each component (orange) fails.
    \textbf{(Right) Illustration of the proposed score-based graph generation framework}. GDSS generates $\bm{X}$ and $\bm{A}$ simultaneously by modeling the dependency through time, whereas GDSS-seq generates them sequentially. EDP-GNN generates only $\bm{A}$ with $\bm{X}$ sampled from training data. Note that GDSS and GDSS-seq are based on a continuous-time diffusion process, while EDP-GNN is based on a discrete-step perturbation procedure.}
    \label{fig:concept}
    \vspace{-0.1in}
\end{figure*}

\vspace{-0.1in}
\paragraph{Score-based Graph Generation}\hspace{-10pt} 
To the best of our knowledge, \citet{score-based/graph/1} is the only work that approaches graph generation with the score-based generative model, which aims to generate graphs by estimating the score of the adjacency matrices at a finite number of noise scales, and using Langevin dynamics to sample from a series of decreasing noise scales. However, Langevin dynamics requires numerous sampling steps for each noise scale, thereby taking a long time for the generation and having difficulty in generating large-scale graphs. Furthermore, \citet{score-based/graph/1} focuses on the generation of adjacency without the generation of node features, resulting in suboptimal learning of the distributions of node-attributed graphs such as molecular graphs. A naive extension of the work of \citet{score-based/graph/1} that generates the node features and the adjacency either simultaneously or alternately will still be suboptimal, since it cannot capture the complex dependency between the nodes and edges. Therefore, we propose a novel score-based generative framework for graphs that can interdependently generate the nodes and edges. Specifically, we propose a novel diffusion process for graphs through a system of SDEs that smoothly transforms the data distribution to known prior and vice versa, which overcomes the limitation of the previous discrete-step perturbation procedure. 

%% file: 4_method.tex
\section{Graph Diffusion via the System of SDEs}
In this section, we introduce our novel continuous-time score-based generative framework for modeling graphs using the system of SDEs. We first explain our proposed graph diffusion process via a system of SDEs in Section~\ref{subsec:diffusion}, then derive new objectives for estimating the gradients of the joint log-density with respect to each component in Section~\ref{subsec:objectives}. Finally, we present an effective method for solving the system of reverse-time SDEs in Section~\ref{subsec:solve}.

\subsection{Graph Diffusion Process \label{subsec:diffusion}}
The goal of graph generation is to synthesize graphs that closely follow the distribution of the observed set of graphs. To bypass the difficulty of directly representing the distribution, we introduce a continuous-time score-based generative framework for the graph generation. Specifically, we propose a novel graph diffusion process via the system of SDEs that transforms the graphs to noise and vice versa, while modeling the dependency between nodes and edges. We begin by explaining the proposed diffusion process for graphs. 

A graph $\bm{G}$ with $N$ nodes is defined by its node features $\bm{X}\!\in\!\mathbb{R}^{\!N\!\times\! F}$ and the weighted adjacency matrix $\bm{A}\!\in\!\mathbb{R}^{\!N\!\times\! N}$ as $\bm{G}\!=\!(\bm{X},\!\bm{A})\!\in\! \mathbb{R}^{\!N\!\times\! F}\!\!\times\mathbb{R}^{\!N\!\times\! N}\!\! \coloneqq\mathcal{G}$, where $F$ is the dimension of the node features. To model the dependency between $\bm{X}$ and $\bm{A}$, we propose a forward diffusion process of graphs that transforms both the node features and the adjacency matrices to a simple noise distribution. Formally, the diffusion process can be represented as the trajectory of random variables $\left\{\bm{G}_t\!=\!(\bm{X}_t,\!\bm{A}_t)\right\}_{t\in[0,\!T]}$ in a fixed time horizon $[0,\!T]$, where $\bm{G}_0$ is a graph from the data distribution $p_{data}$. The diffusion process can be modeled by the following It\^{o} SDE:
\begin{equation}
\fontsize{9.5pt}{9.5pt}\selectfont
\begin{aligned}
    \mathrm{d}\bm{G}_t = \mathbf{f}_t(\bm{G}_t)\mathrm{d}t + \mathbf{g}_t(\bm{G}_t)\mathrm{d}\mathbf{w} , \;\;\; \bm{G}_0\sim p_{data},
    \label{eq:forward_diffusion}
\end{aligned}
\fontsize{10pt}{10pt}\selectfont
\end{equation}
where $\mathbf{f}_t(\cdot)\colon\!\mathcal{G}\! \to\! \mathcal{G}$~\footnote{$t$-subscript represents functions of time: $F_t(\cdot)\!\coloneqq\! F(\cdot,t)$.} 
is the linear drift coefficient, $\mathbf{g}_t(\cdot)\!\colon\! \mathcal{G}\!\to\! \mathcal{G}\!\times\! \mathcal{G}$ is the diffusion coefficient, and $\mathbf{w}$ is the standard Wiener process. 
Intuitively, the forward diffusion process of Eq.~\eqref{eq:forward_diffusion} smoothly transforms both $\bm{X}_0$ and $\bm{A}_0$ by adding infinitesimal noise $\mathrm{d}\mathbf{w}$ at each infinitesimal time step $\mathrm{d}t$. The coefficients of the SDE, $\mathbf{f}_t$ and $\mathbf{g}_t$, are chosen such that at the terminal time horizon $T$, the diffused sample $\bm{G}_T$ approximately follows a prior distribution that has a tractable form to efficiently generate the samples, for example Gaussian distribution. For ease of the presentation, we choose $\mathbf{g}_t(\bm{G}_t)$ to be a scalar function $g_t$. Note that ours is the first work that proposes a diffusion process for generating a whole graph consisting of nodes and edges with attributes, in that the work of \citet{score-based/graph/1} (1) utilizes the finite-step perturbation of multiple noise scales, and (2) only focuses on the perturbation of the adjacency matrices while using the fixed node features sampled from the training data.

In order to generate graphs that follow the data distribution, we start from samples of the prior distribution and traverse the diffusion process of Eq.~\eqref{eq:forward_diffusion} backward in time. Notably, the reverse of the diffusion process in time is also a diffusion process described by the following reverse-time SDE~\cite{reversesde,score-based/2}:
\begin{equation}
\fontsize{9.5pt}{9.5pt}\selectfont
\begin{aligned}
    \mathrm{d}\bm{G}_t = \left[ \mathbf{f}_t(\bm{G}_t) - g_t^2\score{\bm{G}_t} \right] \mathrm{d}\overbar{t} + g_t\mathrm{d}\bar{\mathbf{w}}
    \label{eq:reverse_diffusion},
\end{aligned}
\fontsize{10pt}{10pt}\selectfont
\end{equation}
where $p_t$ denotes the marginal distribution under the forward diffusion process at time $t$, $\bar{\mathbf{w}}$ is a reverse-time standard Wiener process, and $\mathrm{d}\overbar{t}$ is an infinitesimal negative time step. However, solving Eq.~\eqref{eq:reverse_diffusion} directly requires the estimation of high-dimensional score $\score{\bm{G}_t} \!\in\! \mathbb{R}^{N\!\times\! F}\!\!\times\!\mathbb{R}^{N\!\times\! N}$, which is expensive to compute. To bypass this computation, we propose a novel reverse-time diffusion process equivalent to Eq.~\eqref{eq:reverse_diffusion}, modeled by the following system of SDEs: 
\begin{align}
\fontsize{9pt}{9pt}\selectfont
\hspace*{-3mm}
\begin{cases}
    \! \mathrm{d}\bm{X}_t \! = \! \left[\mathbf{f}_{1,t}(\bm{X}_t) - g_{1,t}^2\! \pscore{\bm{X}_t}{\bm{X}_t,\! \bm{A}_t} \right]\! \mathrm{d}\overbar{t} + g_{1,t}\mathrm{d}\bar{\mathbf{w}}_{1} \\[5pt]
    \! \mathrm{d}\bm{A}_t = \! \left[\mathbf{f}_{2,t}(\bm{A}_t) - g_{2,t}^2\! \pscore{\bm{A}_t}{\bm{X}_t,\! \bm{A}_t} \right]\! \mathrm{d}\overbar{t} + g_{2,t}\mathrm{d}\bar{\mathbf{w}}_{2}
\end{cases}\label{eq:system_sdes}
\fontsize{9pt}{9pt}\selectfont
\hspace{-5mm}
\end{align}
where $\mathbf{f}_{1,t}$ and $\mathbf{f}_{2,t}$ are linear drift coefficients satisfying $\mathbf{f}_t(\bm{X},\bm{A}) = \left(\mathbf{f}_{1,t}(\bm{X}), \mathbf{f}_{2,t}(\bm{A}) \right)$, $g_{1,t}$ and $g_{2,t}$ are scalar diffusion coefficients, and $\bar{\mathbf{w}}_{1}$, $\bar{\mathbf{w}}_{2}$ are reverse-time standard Wiener processes. We refer to these forward and reverse diffusion processes of graphs as \emph{Graph Diffusion via the System of SDEs} (GDSS). Notably, each SDE in Eq.~\eqref{eq:system_sdes} describes the diffusion process of each component, $\bm{X}$ and $\bm{A}$, respectively, which presents a new perspective of interpreting the diffusion of a graph as the diffusion of each component that are interrelated through time. In practice, we can choose different types of SDEs for each component-wise diffusion that best suit the generation process.

The key property of GDSS is that the diffusion processes in the system are dependent on each other, related by the gradients of the joint log-density $\pscore{\bm{X}_t}{\bm{X}_t,\! \bm{A}_t}$ and $\pscore{\bm{A}_t}{\bm{X}_t,\! \bm{A}_t}$, which we refer to as the \emph{partial score} functions. By leveraging the partial scores to model the dependency between the components through time, GDSS is able to represent the diffusion process of a whole graph, consisting of nodes and edges. To demonstrate the importance of modeling the dependency, we present two variants of our proposed GDSS and compare their generative performance.

The first variant is the continuous-time version of EDP-GNN~\cite{score-based/graph/1}. By ignoring the diffusion process of $\bm{X}$ in Eq.~\eqref{eq:system_sdes} with $\bm{f}_{1,t}\!\!=\!g_{1,t}\!=\!0$ and choosing the prior distribution of $\bm{X}$ as the data distribution, we obtain a diffusion process of $\bm{A}$ that generalizes the discrete-step noise perturbation procedure of EDP-GNN. Therefore, EDP-GNN can be considered as a special example of GDSS without the diffusion process of the node features, which further replaces the diffusion by the discrete-step perturbation with a finite number of noise scales. We present another variant of GDSS that generates $\bm{X}$ and $\bm{A}$ sequentially instead of generating them simultaneously. By neglecting some part of the dependency through the assumptions $\nabla_{\!\!\bm{A}_t}\!\log p_t(\bm{X}_t,\!\bm{A}_t)\! \approx\! \pscore{\bm{X}_t}{\bm{X}_t}$ and $\nabla_{\!\!\bm{A}_t}\!\log p_t(\bm{X}_t,\!\bm{A}_t)\! \approx\! \nabla_{\!\!\bm{A}_t}\!\log p_t(\bm{X}_0,\!\bm{A}_t)$, we can derive the following SDEs for the diffusion process of the variant:
\begin{equation}
\hspace*{-3mm}
\fontsize{9pt}{9pt}\selectfont
\begin{aligned}
    \mathrm{d}\bm{X}_t \!= &\left[\mathbf{f}_{1,t}(\bm{X}_t) - g_{1,t}^2 \pscore{\bm{X}_t}{\bm{X}_t} \right]\! \mathrm{d}\overbar{t} + g_{1,t}\mathrm{d}\bar{\mathbf{w}}_{1}, \\
    \mathrm{d}\bm{A}_t \!= &\left[\mathbf{f}_{2,t}(\bm{A}_t) - g_{2,t}^2\pscore{\bm{A}_t}{\bm{X}_0,\bm{A}_t} \right]\! \mathrm{d}\overbar{t} + g_{2,t}\mathrm{d}\bar{\mathbf{w}}_{2}, 
\hspace{-2mm}
\end{aligned}
\fontsize{10pt}{10pt}\selectfont
\label{eq:sequential_sdes}
\end{equation}
which are sequential in the sense that the reverse diffusion process of $\bm{A}$ is determined by $\bm{X}_0$, the result of the reverse diffusion of $\bm{X}$. Thus simulating Eq.~\eqref{eq:sequential_sdes} can be interpreted as generating the node features $\bm{X}$ first, then generating the adjacency $\bm{A}$ sequentially, which we refer to as \emph{GDSS-seq}.

The reverse diffusion process of these two variants, EDP-GNN and GDSS-seq, are visualized in Figure~\ref{fig:concept} as the red trajectory in the joint $(\bm{X},\!\bm{A})$-space, where the trajectory is constrained to the hyperplane defined by $\bm{X}_t\!=\!\bm{X}_0$, therefore do not fully reflect the dependency. On the other hand, GDSS represented as the green trajectory is able to diffuse freely from the noise to the data distribution by modeling the joint distribution through the system of SDEs, thereby successfully generating samples from the data distribution.

\begin{figure}[!h]
    \vspace{-0.1in}
    \centering
    \includegraphics[width=\linewidth]{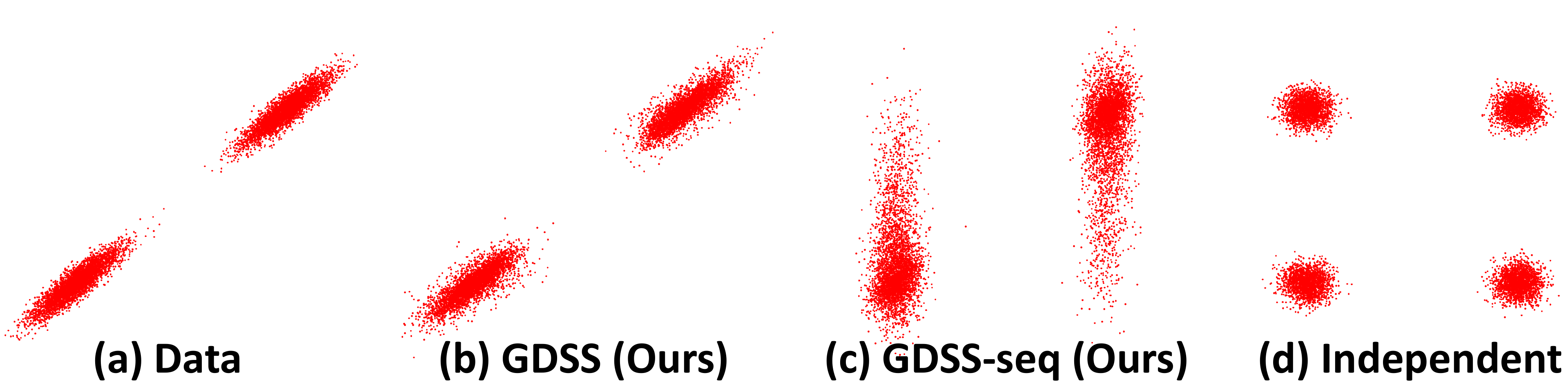}
    \vspace{-0.25in}
    \caption{\small \textbf{A toy experiment on modeling the dependency.} GDSS successfully models the correlation, whereas others fail. See Appendix~\ref{sec:app:toy} for more details of the experiment.
    }\label{fig:toy}
    \vspace{-0.1in}
\end{figure}
To empirically verify this observation, we conduct a simple experiment where the data distribution is a bivariate Gaussian mixture. The generated samples of each process are shown in Figure~\ref{fig:toy}. While GDSS successfully represents the correlation of the two variables, GDSS-seq fails to capture their covariance and generates samples that deviate from the data distribution. We observe that EDP-GNN shows a similar result.
We further extensively validate the effectiveness of our GDSS in modeling the dependency in Section~\ref{sec:experiments}.

Note that once the partial scores in Eq.~\eqref{eq:system_sdes} are known for all $t$, the system of reverse-time SDEs can be used as a generative model by simulating the system backward in time, which we further explain in Section~\ref{subsec:solve}. In order to estimate the partial scores with neural networks, we introduce novel training objectives for GDSS in the following subsection.

\subsection{Estimating the Partial Score Functions \label{subsec:objectives}}

\paragraph{Training Objectives}
The partial score functions can be estimated by training the time-dependent score-based models $\bm{s}_{\theta, t}$ and $\bm{s}_{\phi, t}$, so that $\bm{s}_{\theta, t}(\bm{G}_t)\!\approx\!\pscore{\bm{X}_t}{\bm{G}_t}$ and $\bm{s}_{\phi, t}(\bm{G}_t)\!\approx\!\pscore{\bm{A}_t}{\bm{G}_t}$. However, the objectives introduced in the previous works for estimating the score function are not directly applicable, since the partial score functions are defined as the gradient of each component, not the gradient of the data as in the score function. Thus we derive new objectives for estimating the partial scores. 

Intuitively, the score-based models should be trained to minimize the distance to the corresponding ground-truth partial scores. In order to minimize the Euclidean distance, we introduce new objectives that generalize the score matching~\cite{score-matching, score-based/2} to the estimation of the partial scores for the given graph dataset, as follows:
\begin{equation}
\fontsize{9pt}{9pt}\selectfont
\hspace*{-2mm}
\begin{aligned}
    &\min_{\theta}\mathbb{E}_t\!\! \left\{\!\lambda_1(t) \mathbb{E}_{\bm{G}_0}\mathbb{E}_{\bm{G}_t|\bm{G}_0}\! \left\| \bm{s}_{\theta, t}(\bm{G}_t) - \pscore{\bm{X}_t}{\bm{G}_t} \right\|^2_2\right\} \\
    &\min_{\phi}\mathbb{E}_t\!\! \left\{\!\lambda_2(t) \mathbb{E}_{\bm{G}_0}\mathbb{E}_{\bm{G}_t|\bm{G}_0}\! \left\| \bm{s}_{\phi,t}(\bm{G}_t) - \pscore{\bm{A}_t}{\bm{G}_t} \right\|^2_2\right\},
\hspace*{-2mm}
    \label{eq:score_matching}
\end{aligned}
\fontsize{10pt}{10pt}\selectfont
\end{equation}
where $\lambda_1(t)$ and $\lambda_2(t)$ are positive weighting functions and $t$ is uniformly sampled from $[0,T]$. The expectations are taken over the samples $\bm{G}_0\!\sim\!p_{data}$ and $\bm{G}_t\!\sim\!p_{0t}(\bm{G}_t|\bm{G}_0)$, where $p_{0t}(\bm{G}_t|\bm{G}_0)$ denotes the transition distribution from $p_0$ to $p_t$ induced by the forward diffusion process.

Unfortunately, we cannot train directly with Eq.~\eqref{eq:score_matching} since the ground-truth partial scores are not analytically accessible in general. Therefore we derive tractable objectives equivalent to Eq.~\eqref{eq:score_matching}, by leveraging the idea of denoising score matching~\cite{vincent,score-based/2} to the partial scores, as follows (see Appendix~\ref{sec:app/matching} for the derivation):
\begin{equation*}
\fontsize{8.5pt}{8.5pt}\selectfont
\begin{aligned}
    &\min_{\theta}\mathbb{E}_t\left\{\lambda_1(t)\mathbb{E}_{\bm{G}_0}\mathbb{E}_{\bm{G}_t|\bm{G}_0} \left\|
    \bm{s}_{\theta, t}(\bm{G}_t) - \dscore{X}{G} \right\|^2_2\right\} \\
    &\min_{\phi}\mathbb{E}_t\left\{\lambda_2(t)\mathbb{E}_{\bm{G}_0}\mathbb{E}_{\bm{G}_t|\bm{G}_0} \left\| 
    \bm{s}_{\phi, t}(\bm{G}_t) - \dscore{A}{G} \right\|^2_2\right\}.
\end{aligned}
\end{equation*}
Since the drift coefficient of the forward diffusion process in Eq.~\eqref{eq:forward_diffusion} is linear, the transition distribution $p_{0t}(\bm{G}_t|\bm{G}_0)$ can be separated in terms of $\bm{X}_t$ and $\bm{A}_t$ as follows:
\begin{equation}
\fontsize{9pt}{9pt}\selectfont
\begin{aligned}
    p_{0t}(\bm{G}_t|\bm{G}_0) = p_{0t}(\bm{X}_t|\bm{X}_0)\;p_{0t}(\bm{A}_t|\bm{A}_0).
    \label{eq:trainsition_kernel}
\end{aligned}
\fontsize{10pt}{10pt}\selectfont
\end{equation}
Notably, we can easily sample from the transition distributions of each components, $p_{0t}(\bm{X}_t|\bm{X}_0)$ and $p_{0t}(\bm{A}_t|\bm{A}_0)$, as they are Gaussian distributions where the mean and variance are tractably determined by the coefficients of the forward diffusion process~\citep{kernel_derivation}. From Eq.~\eqref{eq:trainsition_kernel}, we propose new training objectives which are equivalent to Eq.~\eqref{eq:score_matching} (see Appendix~\ref{sec:app/kernel} for the detailed derivation):
\begin{equation}
\hspace*{-3.2mm}
\fontsize{8pt}{8pt}\selectfont
\begin{aligned}
    &\min_{\theta}\mathbb{E}_t\!\! \left\{\! \lambda_1(t) \mathbb{E}_{\bm{G}_0}\! \mathbb{E}_{\bm{G}_t|\bm{G}_0}\!\! \left\| \bm{s}_{\theta,t}(\bm{G}_t) - \dscore{X}{X} \right\|^2_2\right\} \\
    &\min_{\phi}\mathbb{E}_t\!\! \left\{\! \lambda_2(t) \mathbb{E}_{\bm{G}_0}\! \mathbb{E}_{\bm{G}_t|\bm{G}_0}\!\! \left\| \bm{s}_{\phi,t}(\bm{G}_t) - \dscore{A}{A} \right\|^2_2\right\} \label{eq:objective}
\hspace*{-5mm}
\end{aligned}
\fontsize{10pt}{10pt}\selectfont
\end{equation}
The expectations in Eq.~\eqref{eq:objective} can be efficiently computed using the Monte Carlo estimate with the samples $(t,\bm{G}_0,\bm{G}_t)$. Note that estimating the partial scores is not equivalent to estimating $\score{\bm{X}_t}$ or $\score{\bm{A}_t}$, the main objective of previous score-based generative models, since estimating the partial scores requires capturing the dependency between $\bm{X}_t$ and $\bm{A}_t$ determined by the joint probability through time. As we can effectively estimate the partial scores by training the time-dependent score-based models with the objectives of Eq.~\eqref{eq:objective}, what remains is to find the models that can learn the partial scores of the underlying distribution of graphs. Thus we propose new architectures for the score-based models in the next paragraph. 

\paragraph{Permuation-equivariant Score-based Model \label{subsec:objectives:model}} 
Now we propose new architectures for the time-dependent score-based models that can capture the dependencies of $\bm{X}_t$ and $\bm{A}_t$ through time, based on graph neural networks (GNNs). First, we present the score-based model $\bm{s}_{\phi,t}$ to estimate $\nabla_{\!\!\bm{A}_t}\!\log p_t(\bm{X}_t,\!\bm{A}_t)$ which has the same dimensionality as $\bm{A}_t$. We utilize the graph multi-head attention~\cite{GMT} to distinguish important relations between nodes, and further leverage the higher-order adjacency matrices to represent the long-range dependencies as follows:
\begin{equation}
\fontsize{9pt}{9pt}\selectfont
\begin{aligned}
    \bm{s}_{\phi,t}(\bm{G}_t) = \text{MLP}\left(\left[ \left\{ \text{GMH}\left(\bm{H}_i,\bm{A}^p_t\right)\right\}^{K,P}_{i=0,p=1} \right]\right),
\end{aligned}
\fontsize{10pt}{10pt}\selectfont
\end{equation}
where $\bm{A}^{p}_t$ are the higher-order adjacency matrices, $\bm{H}_{i+1}\!=\!\text{GNN}(\bm{H}_i,\!\bm{A}_t)$ with $\bm{H}_0\!=\!\bm{X}_t$ given, $[\cdot]$ denotes the concatenation operation, GMH denotes the graph multi-head attention block, and $K$ denotes the number of GMH layers. We also present the score-based model $\bm{s}_{\theta,t}$ to estimate $\pscore{\bm{X}_t}{\bm{X}_t,\bm{A}_t}$ which has the same dimensionality as $\bm{X}_t$, where we use multiple layers of GNNs to learn the partial scores from the node representations as follows:
\begin{equation}
\fontsize{9pt}{9pt}\selectfont
\begin{aligned}
    \bm{s}_{\theta,t}(\bm{G}_t) = \text{MLP}([\left\{\bm{H}_i\right\}^{L}_{i=0}]),
\end{aligned}
\fontsize{10pt}{10pt}\selectfont
\end{equation}
where $\bm{H}_{i+1}\!=\!\text{GNN}(\bm{H}_i,\!\bm{A}_t)$ with $\bm{H}_0\!=\!\bm{X}_t$ given and $L$ denotes the number of GNN layers. Here GMH layers can be used instead of simple GNN layers with additional computation costs, which we analyze further in Section~\ref{subsec:ablation}. The architectures of the score-based models are illustrated in Figure~\ref{fig:model_architecture} of Appendix. Moreover, following \citet{score-based/1}, we incorporate the time information to the score-based models by scaling the output of the models with the standard deviation of the transition distribution at time $t$.

Note that since the message-passing operations of GNNs and the attention function used in GMH are permutation-equivariant~\cite{permutation_equivariance}, the proposed score-based models are also equivariant, and theryby from the result of \citet{score-based/graph/1}, the log-likelihood implicitly defined by the models is guaranteed to be permutation-invariant.

\subsection{Solving the System of Reverse-time SDEs~\label{subsec:solve}}
In order to use the reverse-time diffusion process as a generative model, it requires simulating the system of reverse-time SDEs in Eq.~\eqref{eq:system_sdes}, which can be approximated using the trained score-based models $\bm{s}_{\theta,t}$ and $\bm{s}_{\phi,t}$ as follows:
\begin{equation}
\fontsize{9pt}{9pt}\selectfont
\begin{aligned}
    \begin{cases}
        \mathrm{d}\bm{X}_t \!= \eqmakebox[F]{$\mathbf{f}_{1,t}(\bm{X}_t) \mathrm{d}\overbar{t} + g_{1,t}\mathrm{d}\mathbf{\bar{w}}_1$} \; \eqmakebox[S]{$- g^2_{1,t}\bm{s}_{\theta,t}(\bm{X}_t,\!\bm{A}_t)\mathrm{d}\overbar{t}$} \\
        \mathrm{d}\bm{A}_t\, \!= \eqmakebox[LHS]{$\mathbf{f}_{2,t}(\bm{A}_t) \mathrm{d}\overbar{t} + g_{2,t}\mathrm{d}\mathbf{\bar{w}}_2$} \; \eqmakebox[S]{$- g^2_{2,t}\bm{s}_{\!\phi,t}(\bm{X}_t,\!\bm{A}_t)\mathrm{d}\overbar{t}$}
    \end{cases} \\[-1.2\normalbaselineskip]
    \underbrace{\eqmakebox[F]{\mathstrut}}_{F}
    \phantom{}
    \underbrace{\eqmakebox[S]{\mathstrut}}_{S}
    \phantom{a\;}
    \label{eq:estimated_system_sdes}
\end{aligned}
\fontsize{10pt}{10pt}\selectfont
\end{equation}
However, solving the system of two diffusion processes that are interdependently tied by the partial scores brings about another difficulty. Thus we propose a novel integrator, \emph{Symmetric Splitting for System of SDEs} (S4) to simulate the system of reverse-time SDEs, that is efficient yet accurate, inspired by the Symmetric Splitting CLD Sampler (SSCS)~\cite{CLD} and the Predictor-Corrector Sampler (PC sampler)~\cite{score-based/2}.

Specifically, at each discretized time step $t$, S4 solver consists of three steps: the score computation, the correction, and the prediction. First, S4 computes the estimation of the partial scores with respect to the predicted $\bm{G}_{t}$, using the score-based models $\bm{s}_{\theta,t}$ and $\bm{s}_{\phi,t}$, where the computed partial scores are later used for both the correction and the prediction steps. After the score computation, we perform the correction step by leveraging a score-based MCMC method, for example Langevin MCMC~\cite{LangevinMCMC}, in order to obtain calibrated sample $\bm{G}'_t$ from $\bm{G}_{t}$. Here we exploit the precomputed partial scores for the score-based MCMC approach. Then what remains is to predict the solution for the next time step $t-\delta t$, going backward in time. 

The prediction of the state at time $t'$ follows the marginal distribution $p_{t'}$ described by the Fokker-Planck equation induced by Eq.~\eqref{eq:estimated_system_sdes}, where the Fokker-Planck operators $\fop_F$ and $\fop_S$ correspond to the $F$-term and $S$-term, respectively~\footnote{Details of the Fokker-Planck operators are given in Appendix~\ref{sec:app:solve}.}. Inspired by \citet{CLD}, we formalize an intractable solution to Eq.~\eqref{eq:estimated_system_sdes} with the classical propagator $e^{t(\fop_F+\fop_S)}$ which gives light to finding efficient prediction method. From the result of the symmetric Trotter theorem~\cite{Trotter,Strang}, the propagation of the calibrated state $\bm{G}'_t$ from time $t$ to $t-\delta t$ following the dynamics of Eq.~\eqref{eq:estimated_system_sdes}
can be approximated by applying $e^{\frac{\delta t}{2}\fop_F} e^{\delta t\fop_S} e^{\frac{\delta t}{2}\fop_F}$ to $\bm{G}'_t$. Observing the operators individually, the action of first $e^{\frac{\delta t}{2}\fop_F}$ describes the dynamics of the $F$-term in Eq.~\eqref{eq:estimated_system_sdes} from time $t$ to $t-\delta t/2$, which is equal to sampling from the transition distribution of the forward diffusion in Eq.~\eqref{eq:forward_diffusion} as follows (see Appendix~\ref{sec:app:s4_derivation} and ~\ref{sec:app:transition_kernel} for the derivation of the action and the transition distribution.):
\begin{equation}
\fontsize{9.5pt}{9.5pt}\selectfont
\begin{aligned}
    e^{\frac{\delta t}{2}\fop_F}\bm{G} = \tilde{\bm{G}}\sim p_{t,t-\delta t/2}(\tilde{\bm{G}}|\bm{G}).
\end{aligned}
\fontsize{10pt}{10pt}\selectfont
\end{equation}
On the other hand, the action of $e^{\delta t\fop_S}$ is not analytically accessible, so we approximate the action with a simple Euler method (EM) that solves the ODE corresponding to the $S$-term in Eq.~\eqref{eq:estimated_system_sdes}. Here we use the precomputed partial scores again, which is justified due to the action of the first $e^{\frac{\delta t}{2}\fop_{F}}$ on a sufficiently small half-step $\delta t/2$. Lastly, the action of remaining $e^{\frac{\delta t}{2}\fop_F}$ corresponds to sampling from the transition distribution from time $t-\delta t/2$ to $t-\delta t$, which results in the approximated solution $\bm{G}_{t-\delta t}$. We provide the pseudo-code for the S4 solver in Algorithm~\ref{alg:symmetric_splitting} of Appendix. To obtain more accurate solution with additional cost of computation, one might consider using a higher-order integrator such as Runge-Kutta method to approximate the action of the operator $e^{\delta t\fop_S}$, and further leverage HMC~\cite{HMC} for the correction step instead of Langevin MCMC. 

Note that although S4 and PC sampler both carry out the prediction and correction steps, 
S4 solver is far more efficient in terms of computation since compared to the PC sampler, S4 requires half the number of forward passes to the score-based models which dominates the computational cost of solving the SDEs. Moreover, the proposed S4 solver can be used to solve a general system of SDEs, including the system with mixed types of SDEs such as those of Variance Exploding (VE) SDE and Variance Preserving (VP) SDE~\cite{score-based/2}, whereas SSCS is limited to solving a specific type of SDE, namely CLD.

%% file: 5_experiments.tex
\input{table/mmd}
\begin{figure*}[ht]
\vspace{-0.1in}
\begin{minipage}{0.3\linewidth}
    \centering
    \includegraphics[width=1\linewidth]{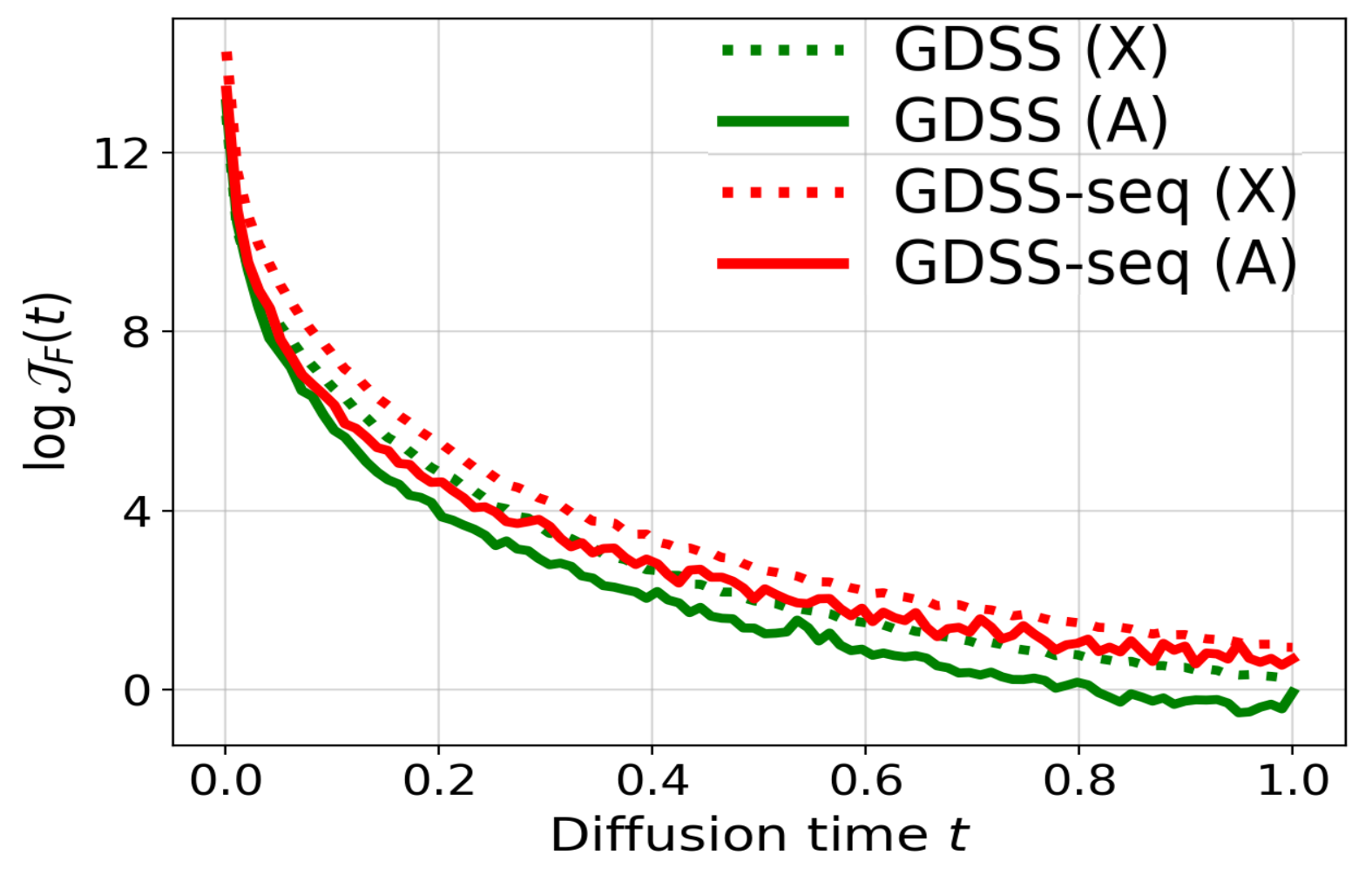}
\end{minipage}
\hfill
\begin{minipage}{0.66\linewidth}
    \vspace{-0.1in}
    \input{table/solver}
\end{minipage}
    \vspace{-0.15in}
    \caption{\small \textbf{(Left) Complexity of the score-based models} measured by the Frobenius norm of the Jacobian of the model. We compare GDSS (green) against GDSS-seq (red), where the solid and dotted lines denote the models estimating the partial scores with respect to $\bm{X}$ and $\bm{A}$, respectively. \textbf{(Right) Comparison between fixed step size SDE solvers.} We measure the time for the generation of 128 graphs.}
    \label{fig:vis}
    \vspace{-0.1in}
\end{figure*} 

\section{Experiments \label{sec:experiments}}
We experimentally validate the performance of our method in generation of generic graphs as well as molecular graphs.

\subsection{Generic Graph Generation~\label{exp:generic_graph}}
To verify that GDSS is able to generate graphs that follow the underlying data distribution, we evaluate our method on generic graph generation tasks with various datasets.

\paragraph{Experimental Setup}
We first validate GDSS by evaluating the quality of the generated samples on four generic graph datasets, including synthetic and real-world graphs with varying sizes: (1) Ego-small, 200 small ego graphs drawn from larger Citeseer network dataset~\cite{citeseer}, (2) Community-small, 100 randomly generated community graphs, (3) Enzymes, 587 protein graphs which represent the protein tertiary structures of the enzymes from the BRENDA database~\cite{enzymes}, and (4) Grid, 100 standard 2D grid graphs. For a fair comparison, we follow the experimental and evaluation setting of \citet{you2018graphrnn} with the same train/test split. We use the maximum mean discrepancy (MMD) to compare the distributions of graph statistics between the same number of generated and test graphs. Following \citet{you2018graphrnn}, we measure the distributions of degree, clustering coefficient, and the number of occurrences of orbits with 4 nodes. Note that we use the Gaussian Earth Mover's Distance (EMD) kernel to compute the MMDs instead of the total variation (TV) distance used in \citet{liao2019efficient}, since the TV distance leads to an indefinite kernel and an undefined behavior~\cite{kernel}. Please see Appendix~\ref{app:sec:generic_graph} for more details.

\input{table/qm9+zinc250k}
\begin{figure*}[ht]
    \vspace{-0.1in}
    \centering
    \includegraphics[width=\linewidth]{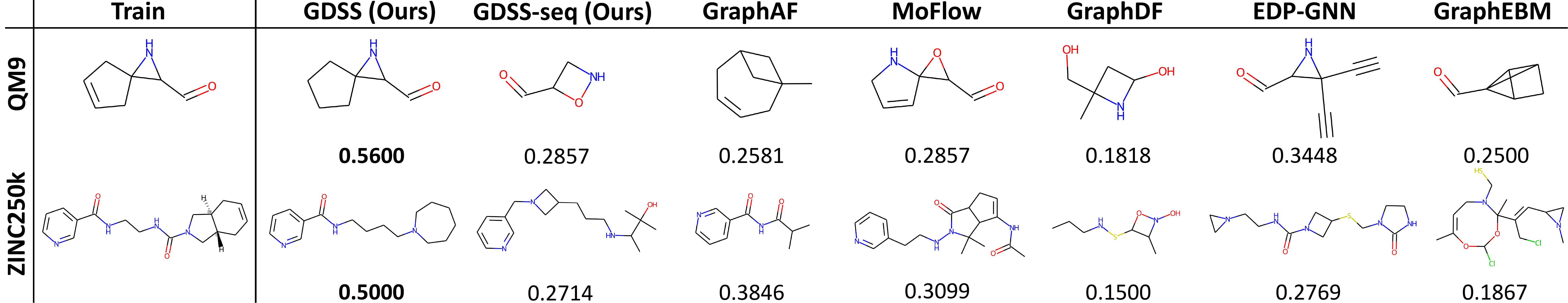}
    \vspace{-0.25in}
    \caption{\small \textbf{Visualization of the generated molecules with maximum Tanimoto similarity} to the molecule from the dataset. The top row shows QM9 molecules while the bottom row shows ZINC250k molecules. For each generated molecule, we display the similarity value at the bottom. The pairwise Tanimoto similarity is calculated based on the standard Morgan fingerprints with radius 2 and 1024 bits.}
    \label{fig:mols}
    \vspace{-0.1in}
\end{figure*}

\paragraph{Implementation Details and Baselines}
We compare our proposed method against the following deep generative models. \textbf{GraphVAE}~\cite{drug-design/1} is a one-shot VAE-based model. \textbf{DeepGMG}~\cite{deepgmg} and \textbf{GraphRNN}~\cite{you2018graphrnn} are autoregressive RNN-based models. \textbf{GNF}~\cite{GNF} is a one-shot flow-based model. \textbf{GraphAF}~\cite{shi2020graphaf} is an autoregressive flow-based model. \textbf{EDP-GNN}~\cite{score-based/graph/1} is a one-shot score-based model. \textbf{GraphDF}~\cite{luo2021graphdf} is an autoregressive flow-based model that utilizes discrete latent variables. For GDSS, we consider three types of SDEs introduced by \citet{score-based/2}, namely VESDE, VPSDE, and sub-VP SDE, for the diffusion processes of each component, and either use the PC sampler or the S4 solver to solve the system of SDEs. We provide further implementation details of the baselines and our GDSS in Appendix~\ref{app:sec:generic_graph}.

\paragraph{Results}
Table~\ref{tab:mmd} shows that the proposed GDSS significantly outperforms all the baseline one-shot generative models including EDP-GNN, and also outperforms the autoregressive baselines on most of the datasets, except Grid. Moreover, GDSS shows competitive performance to the state-of-the-art autoregressive model GraphRNN on generating large graphs, i.e. Grid dataset, for which GDSS is the only method that achieves similar performance. We observe that EDP-GNN, a previous score-based generation model for graphs, completely fails in generating large graphs. The MMD results demonstrate that GDSS can effectively capture the local characteristics of the graphs, which is possible due to modeling the dependency between nodes and edges. We visualize the generated graphs of GDSS in Appendix~\ref{sec/app/vis_generic}.

\paragraph{Complexity of Learning Partial Scores}\hspace{-10pt} 
Learning the partial score $\pscore{\bm{A}_t}{\bm{X}_t,\!\bm{A}_t}$ regarding both $\bm{X}_t$ and $\bm{A}_t$ may seem difficult, compared to learning $\pscore{\bm{A}_t}{\bm{X}_0,\!\bm{A}_t}$ that concerns only $\bm{A}_t$. We empirically demonstrate this is not the case, by measuring the complexity of the score-based models that estimate the partial scores. The complexity of the models can be measured via the squared Frobenius norm of their Jacobians $\mathcal{J}_F(t)$, and we compare the models of GDSS and GDSS-seq trained with the Ego-small dataset. As shown in Figure~\ref{fig:vis}, the complexity of GDSS estimating $\pscore{\bm{A}_t}{\bm{X}_t,\!\bm{A}_t}$ is significantly smaller compared to the complexity of GDSS-seq estimating $\pscore{\bm{A}_t}{\bm{X}_0,\!\bm{A}_t}$, and similarly for $\bm{X}_t$. The result stresses the importance of modeling the node-edge dependency, since whether to model the dependency is the only difference between GDSS and GDSS-seq. Moreover, the generation result of GDSS compared to GDSS-seq in Table~\ref{tab:mmd} verifies that the reduced complexity enables the effective generation of larger graphs.

\subsection{Molecule Generation \label{molecule_generation}}
To show that GDSS is able to capture the complex dependency between nodes and edges, we further evaluate our method for molecule generation tasks.

\paragraph{Experimental Setup} 
We use two molecular datasets, QM9~\cite{ramakrishnan2014quantum} and ZINC250k~\cite{irwin2012zinc}, where we provide the statistics in Table~\ref{tab:data} in Appendix. Following previous works~\citep{shi2020graphaf,luo2021graphdf}, the molecules are kekulized by the RDKit library~\cite{landrum2016rdkit} with hydrogen atoms removed. We evaluate the quality of the 10,000 generated molecules with the following metrics. \textbf{Fr\'{e}chet ChemNet Distance (FCD)}~\cite{preuer2018frechet} evaluates the distance between the training and generated sets using the activations of the penultimate layer of the ChemNet. \textbf{Neighborhood subgraph pairwise distance kernel (NSPDK) MMD}~\cite{nspdk} is the MMD between the generated molecules and test molecules which takes into account both the node and edge features for evaluation. Note that FCD and NSPDK MMD are salient metrics that assess the ability to learn the distribution of the training molecules, measuring how close the generated molecules lie to the distribution. Specifically, FCD measures the ability in the view of molecules in chemical space, while NSPDK MMD measures the ability in the view of the graph structure. \textbf{Validity w/o correction} is the fraction of valid molecules without valency correction or edge resampling, which is different from the metric used in~\citet{shi2020graphaf} and \citet{luo2021graphdf}, since we allow atoms to have formal charge when checking their valency following~\citet{zang2020moflow}, which is more reasonable due to the existence of formal charge in the training molecules. \textbf{Time} measures the time for generating 10,000 molecules in the form of RDKit molecules. We provide further details about the experimental settings, including the hyperparameter search in Appendix~\ref{sec:app/mol/details}.

\paragraph{Baselines}
We compare our GDSS against the following baselines. \textbf{GraphAF}~\cite{shi2020graphaf} is an autoregressive flow-based model. \textbf{MoFlow}~\cite{zang2020moflow} is a one-shot flow-based model. \textbf{GraphDF}~\cite{luo2021graphdf} is an autoregressive flow-based model using discrete latent variables. \textbf{GraphEBM}~\cite{liu2021graphebm} is a one-shot energy-based model that generates molecules by minimizing energies with Langevin dynamics. We also construct modified versions of GraphAF and GraphDF that consider formal charge (\textbf{GraphAF+FC} and \textbf{GraphDF+FC}) for a fair comparison with the baselines and ours. For GDSS, the choice of the diffusion process and the solver is identical to that of the generic graph generation tasks. We provide further details of the baselines and GDSS in Appendix~\ref{sec:app/mol/details}.

\begin{figure*}[th!]
    \centering
    \begin{minipage}{0.49\linewidth}
    \captionof{table}{\small \textbf{Generation results of EDP-GNN using GMH.}} \label{tab:edpgnn_gmh}
    \vspace{-0.1in}
        \input{table/EDPGNN+GMH}
    \end{minipage}
    \hfill
    \begin{minipage}{0.49\linewidth}
    \captionof{table}{\small \textbf{Generation results of the variants of GDSS.}} \label{tab:gdss_variants}
    \vspace{-0.1in}
        \input{table/GDSS_variants}
    \end{minipage} 
\vspace{-0.1in}
\end{figure*}

\vspace{-0.05in}
\paragraph{Results}
As shown in Table~\ref{tab:qm9+zinc250k}, GDSS achieves the highest validity when the post-hoc valency correction is disabled, demonstrating that GDSS is able to proficiently learn the chemical valency rule that requires capturing the node-edge relationship. Moreover, GDSS significantly outperforms all the baselines in NSPDK MMD and most of the baselines in FCD, showing that the generated molecules of GDSS lie close to the data distribution both in the space of graphs and the chemical space. The superior performance of GDSS on molecule generation tasks verifies the effectiveness of our method for learning the underlying distribution of graphs with multiple node and edge types.
We further visualize the generated molecules in Figure~\ref{fig:mols} and Figure~\ref{fig:mols_app} of Appendix~\ref{sec/app/vis_mol}, which demonstrate that GDSS is capable of generating molecules that share a large substructure with the molecules in the training set, whereas the generated molecules of the baselines share a smaller structural portion or even completely differ from the training molecules. 

\vspace{-0.05in}
\paragraph{Time Efficiency}
To validate the practicality of GDSS, we compare the inference time for generating molecules with the baselines. As shown in Table~\ref{tab:qm9+zinc250k}, GDSS not only outperforms the autoregressive models in terms of the generation quality, but also in terms of time efficiency showing $450\times$ speed up on QM9 datasets compared to GraphDF. Moreover, GDSS and GDSS-seq require significantly smaller generation time compared to EDP-GNN, showing that modeling the transformation of graphs to noise and vice-versa as a continuous-time diffusion process is far more efficient than the discrete-step noise perturbation used in EDP-GNN.

\subsection{Ablation Studies \label{subsec:ablation}}
We provide an extensive analysis of the proposed GDSS framework from three different perspectives: (1) The necessity of modeling the dependency between $\bm{X}$ and $\bm{A}$, (2) effectiveness of S4 compared to other solvers, and (3) further comparison with the variants of EDP-GNN and GDSS.

\vspace{-0.05in}
\paragraph{Necessity of Dependency Modeling}\hspace{-10pt} 
To validate that modeling the node-edge dependency is crucial for graph generation, we compare our proposed methods GDSS and GDSS-seq, since the only difference is that the latter only models the dependency of $\bm{A}$ on $\bm{X}$  (Eq.~\eqref{eq:system_sdes} and Eq.~\eqref{eq:sequential_sdes}). From the results in Table~\ref{tab:mmd} and Table~\ref{tab:qm9+zinc250k}, we observe that GDSS constantly outperforms GDSS-seq in all metrics, which proves that accurately learning the distributions of graphs requires modeling the dependency. Moreover, for molecule generation, the node-edge dependency can be directly measured in terms of validity, and the results verify the effectiveness of GDSS modeling the dependency via the system of SDEs.

\vspace{-0.05in}
\paragraph{Significance of S4 Solver}
To validate the effectiveness of the proposed S4 solver, we compare its performance against the non-adaptive stepsize solvers, namely EM and Reverse sampler which are predictor-only methods, and the PC samplers using Langevin MCMC. As shown in the table of Figure~\ref{fig:vis}, S4 significantly outperforms the predictor-only methods, and further outperforms the PC samplers with half the computation time, due to fewer evaluations of the score-based models. We provide more results in Appendix~\ref{sec:app:ablation}.

\vspace{-0.05in}
\paragraph{Variants of EDP-GNN and GDSS}
First, to comprehensively compare EDP-GNN with GDSS, we evaluate the performance of EDP-GNN with GMH layers instead of simple GNN layers. Table~\ref{tab:edpgnn_gmh} shows that using GMH does not necessarily increase the generation quality, and is still significantly outperformed by our GDSS. Moreover, to verify that the continuous-time diffusion process of GDSS is essential, we compare the performance of the GDSS variants. Table~\ref{tab:gdss_variants} shows that GDSS-discrete, which is our GDSS using discrete-step perturbation as in EDP-GNN instead of the diffusion process, performs poorly on molecule generation tasks with an increased generation time, which reaffirms the significance of the proposed diffusion process for graphs. Furthermore, using GMH instead of GNN for the score-based model $\bm{s}_{\theta,t}$ shows comparable results with GDSS.

%% file: table/mmd.tex
\begin{table*}[ht]
    \caption{\small \textbf{Generation results on the generic graph datasets.} We report the MMD distances between the test datasets and generated graphs. Best results are highlighted in bold (smaller the better). 
    The results of the baselines for Ego-small and Community-small dataset are taken from \citet{score-based/graph/1} and \citet{luo2021graphdf}. Hyphen (-) denotes out-of-resources that take more than 10 days or not applicable due to the memory issue. $^*$ denotes our own implementation and $^{\dagger}$ indicates unreproducible results. Due to the space limitation, we provide the standard deviations in Appendix~\ref{sec:app:exp_generic_graph}.}
    \label{tab:mmd}
\vspace{-0.1in}
    \centering
    \resizebox{\textwidth}{!}{
    \renewcommand{\arraystretch}{1.1}
    \renewcommand{\tabcolsep}{8pt}
    \begin{tabular}{l l c c c a c c c a c c c a c c c a}
    \toprule
        & & 
        \multicolumn{4}{c}{{Ego-small}} &
        \multicolumn{4}{c}{{Community-small}} &
        \multicolumn{4}{c}{{Enzymes}} &
        \multicolumn{4}{c}{{Grid}}\\
    \cmidrule(l{2pt}r{2pt}){3-6}    
    \cmidrule(l{2pt}r{2pt}){7-10}
    \cmidrule(l{2pt}r{2pt}){11-14}
    \cmidrule(l{2pt}r{2pt}){15-18}
        & &
        \multicolumn{4}{c}{Real, $4\leq|V|\leq18$} &
        \multicolumn{4}{c}{Synthetic, $12\leq|V|\leq20$} &
        \multicolumn{4}{c}{Real, $10\leq|V|\leq125$} &
        \multicolumn{4}{c}{Synthetic, $100\leq|V|\leq400$} \\
    \cmidrule(l{2pt}r{2pt}){3-6}
    \cmidrule(l{2pt}r{2pt}){7-10}
    \cmidrule(l{2pt}r{2pt}){11-14}
    \cmidrule(l{2pt}r{2pt}){15-18}
        &  & Deg. & Clus. & Orbit & Avg. & Deg. & Clus. & Orbit & Avg. & Deg. & Clus. & Orbit & Avg. & Deg. & Clus. & Orbit & Avg. \\
    \midrule
        \multirow{4}{*}{Autoreg.}
        & DeepGMG & 0.040 & 0.100 & 0.020 & 0.053 & 0.220 & 0.950 & 0.400 & 0.523 & - & - & - & - & - & - & - & - \\
        & GraphRNN & 0.090 & 0.220 & 0.003 & 0.104 & 0.080 & 0.120 & 0.040 & 0.080 &  \textbf{0.017} & 0.062 & 0.046 & 0.042 & \textbf{0.064} & 0.043 & \textbf{0.021} & \textbf{0.043} \\
        & GraphAF$^*$ & 0.03 & 0.11 & \textbf{0.001} & 0.047 & 0.18 & 0.20 & 0.02 & 0.133 & 1.669 & 1.283 & 0.266 & 1.073 & - & - & - & - \\
        & GraphDF$^*$ & 0.04 & 0.13 & 0.01 & 0.060 & 0.06 & 0.12 & 0.03 & 0.070 & 1.503 & 1.061 & 0.202 & 0.922 & - & - & - & - \\
    \midrule
        \multirow{5.5}{*}{One-shot}
        & GraphVAE$^*$ & 0.130 & 0.170 & 0.050 & 0.117 & 0.350 & 0.980 & 0.540 & 0.623 & 1.369 & 0.629 & 0.191 & 0.730 & 1.619 & \textbf{0.0} & 0.919 & 0.846 \\
        & GNF$^{\dagger}$ & 0.030 & 0.100 & \textbf{0.001} & 0.044 & 0.200 & 0.200 & 0.110 & 0.170 & - & - & - & - & - & - & - & - \\
        & EDP-GNN & 0.052 & 0.093 & 0.007 & 0.051 & 0.053 & 0.144 & 0.026 & 0.074 & 0.023 & 0.268 & 0.082 & 0.124 & 0.455 & 0.238 & 0.328 & 0.340 \\
    \cmidrule(l{4pt}r{2pt}){2-18}
        & \textbf{GDSS-seq} (Ours) & 0.032 & 0.027 & 0.011 & 0.023 & 0.090 & 0.123 & \textbf{0.007} & 0.073 & 0.099 & 0.225 & 0.010 & 0.111 & 0.171 & 0.011 & 0.223 & 0.135 \\
        & \textbf{GDSS} (Ours) & \textbf{0.021} & \textbf{0.024} & 0.007 & \textbf{0.017} & \textbf{0.045} & \textbf{0.086} & \textbf{0.007} & \textbf{0.046} & 0.026 & \textbf{0.061} & \textbf{0.009} & \textbf{0.032} & 0.111 & 0.005 & 0.070 & 0.062 \\
    \bottomrule
    \end{tabular}}
\end{table*}

%% file: table/solver.tex
\vspace{0.05in}
\centering
\resizebox{\textwidth}{!}{
\renewcommand{\arraystretch}{1.2}
\renewcommand{\tabcolsep}{7pt}
\begin{tabular}{l cccca cccca}
\toprule
    & \multicolumn{5}{c}{{Community-small}} &
    \multicolumn{5}{c}{{Enzymes}} \\
\cmidrule(l{2pt}r{2pt}){2-6}
\cmidrule(l{2pt}r{2pt}){7-11}
    Solver & Deg. & Clus. & Orbit & Avg. & Time (s) & Deg. & Clus. & Orbit & Avg. & Time (s) \\
\midrule
    EM & 0.055 & 0.133 & 0.017 & 0.068 & \textbf{29.64} & 0.060 & 0.581 & 0.120 & 0.254 & \textbf{153.58} \\
    Reverse & 0.058 & 0.125 & 0.016 & 0.066 & 29.75 & 0.057 & 0.550 & 0.112 & 0.240 & 155.06 \\
    \midrule
    EM + Langevin & 0.045 & \textbf{0.086} & \textbf{0.007} & \textbf{0.046} & 59.93 & 0.028 & 0.062 & 0.010 & 0.033 & 308.42 \\
    Rev. + Langevin & 0.045 & \textbf{0.086} & \textbf{0.007} & \textbf{0.046} & 59.40 & 0.028 & 0.064 & \textbf{0.009} & 0.034 & 310.35 \\
    \midrule
    \textbf{S4} (Ours) & \textbf{0.042} & 0.101 & \textbf{0.007} & 0.050 & 30.93 & \textbf{0.026} & \textbf{0.061} & \textbf{0.009} & \textbf{0.032} & 157.57 \\
\bottomrule
\end{tabular}}

%% file: table/qm9+zinc250k.tex
\begin{table*}[ht]
    \caption{\small \textbf{Generation results on the QM9 and ZINC250k dataset.} Results are the means of 3 different runs, and the best results are highlighted in bold. Values denoted by * are taken from the respective original papers. Other results are obtained by running open-source codes. Val. w/o corr. denotes the Validity w/o correction metric, and values that do not exceed 50\% are underlined. Due to the space limitation, we provide the results of validity, uniqueness, and novelty as well as the standard deviations in Appendix~\ref{sec:app:exp_mol}.}
    \vspace{-0.1in}
    \centering
    \resizebox{\textwidth}{!}{
    \renewcommand{\arraystretch}{1.1}
    \renewcommand{\tabcolsep}{8pt}
    \begin{tabular}{llcccccccc}
    \toprule
    & & \multicolumn{4}{c}{QM9} & \multicolumn{4}{c}{ZINC250k} \\
    \cmidrule(l{2pt}r{2pt}){3-6}
    \cmidrule(l{2pt}r{2pt}){7-10}
        & Method & Val. w/o corr. (\%)$\uparrow$ & NSPDK$\downarrow$ & FCD$\downarrow$ & Time (s)$\downarrow$ & Val. w/o corr. (\%)$\uparrow$ & NSPDK$\downarrow$ & FCD$\downarrow$ & Time (s)$\downarrow$ \\
    \midrule
        \multirow{4}{*}{Autoreg.}
        & GraphAF~\cite{shi2020graphaf} & 67* & 0.020 & 5.268 & 2.52$e^{3}$ & 68* & 0.044 & 16.289 & 5.80$e^{3}$ \\
        & GraphAF+FC & 74.43 & 0.021 & 5.625 & 2.55$e^{3}$ & 68.47 & 0.044 & 16.023 & 6.02$e^{3}$ \\
        & GraphDF~\cite{luo2021graphdf} & 82.67* & 0.063 & 10.816 & 5.35$e^{4}$ & 89.03* & 0.176 & 34.202 & 6.03$e^{4}$ \\
        & GraphDF+FC & 93.88 & 0.064 & 10.928 & 4.91$e^{4}$ & 90.61 & 0.177 & 33.546 & 5.54$e^{4}$ \\
    \midrule
        \multirow{5.5}{*}{One-shot}
        & MoFlow~\cite{zang2020moflow} & 91.36 & 0.017 & 4.467 & \textbf{4.60} & 63.11 & 0.046 & 20.931 & \textbf{2.45$\mathbf{e^{1}}$} \\
        & EDP-GNN~\cite{score-based/graph/1} & \underline{47.52} & 0.005 & \textbf{2.680} & 4.40$e^{3}$ & 82.97 & 0.049 & 16.737 & 9.09$e^{3}$ \\
        & GraphEBM~\cite{liu2021graphebm} & \underline{8.22} & 0.030 & 6.143 & 3.71$e^{1}$ & \underline{5.29} & 0.212 & 35.471 & 5.46$e^{1}$ \\
    \cmidrule(l{2pt}r{2pt}){2-10}
        & \textbf{GDSS-seq} (Ours) & 94.47 & 0.010 & 4.004 & 1.13$e^{2}$ & 92.39 & 0.030 & 16.847 & 2.02$e^{3}$ \\
        & \textbf{GDSS} (Ours) & \textbf{95.72} & \textbf{0.003} & 2.900 & 1.14$e^{2}$ & \textbf{97.01} & \textbf{0.019} & \textbf{14.656} & 2.02$e^{3}$ \\
    \bottomrule
    \end{tabular}}
    \label{tab:qm9+zinc250k}
\end{table*}

%% file: table/EDPGNN+GMH.tex
\centering
\resizebox{\linewidth}{!}{
\renewcommand{\arraystretch}{1.0}
\renewcommand{\tabcolsep}{7pt}
\begin{tabular}{l c c c c c c}
\toprule
    & \multicolumn{3}{c}{Community-small} & \multicolumn{3}{c}{Enzymes} \\
\cmidrule(l{2pt}r{2pt}){2-4}
\cmidrule(l{2pt}r{2pt}){5-7}
    Method & Deg. & Clus. & Orbit & Deg. & Clus. & Orbit \\
\midrule
    EDP-GNN &  0.053 & 0.144 & 0.026 & \textbf{0.023} & 0.268 & 0.082 \\
    EDP-GNN w/ GMH & \textbf{0.033} & 0.130 & 0.035 & 0.047 & 0.328 & 0.051 \\
\midrule
    GDSS (Ours) & 0.045 & \textbf{0.086} & \textbf{0.007} & 0.026 & \textbf{0.061} & \textbf{0.009} \\
\bottomrule
\end{tabular}}

%% file: table/GDSS_variants.tex
\centering
\resizebox{\linewidth}{!}{
\renewcommand{\arraystretch}{0.95}
\renewcommand{\tabcolsep}{8pt}
\hspace{-0.15in}
\begin{tabular}{l c c c c}
\toprule
    & \multicolumn{4}{c}{ZINC250k} \\
    \cmidrule(l{2pt}r{2pt}){2-5}
    Method & Val. w/o corr. (\%) & NSPDK & FCD & Time (s) \\
\midrule
    GDSS-discrete & 53.21 & 0.045 & 22.925 & 6.07$e^{3}$ \\
    GDSS w/ GMH in $\bm{s}_{\theta,t}$ & 94.39 & \textbf{0.015} & \textbf{12.388} & 2.44$e^{3}$ \\
\midrule
    GDSS & \textbf{97.01} & 0.019 & 14.656 & \textbf{2.02$e^{3}$} \\
\bottomrule
\end{tabular}}

%% file: 6_conclusion.tex
\section{Conclusion}
We presented a novel score-based generative framework for learning the underlying distribution of the graphs, which overcomes the limitations of previous graph generative methods. Specifically, we proposed a novel \emph{graph diffusion process via the system of SDEs} (GDSS) that transforms both the node features and adjacency to noise and vice-versa, modeling the dependency between them. Further, we derived new training objectives to estimate the gradients of the joint log-density with respect to each component, and presented a novel integrator to efficiently solve the system of SDEs describing the reverse diffusion process. We validated GDSS on the generation of diverse synthetic and real-world graphs including molecules, on which ours outperforms existing generative methods. We pointed out that modeling the dependency between nodes and edges is crucial for learning the distribution of graphs and shed new light on the effectiveness of score-based generative methods for graphs.

\paragraph{Acknowledgements}
This work was supported by Institute of Information \& communications Technology Planning \& Evaluation (IITP) grant funded by the Korea government(MSIT) (No. 2021-0-02068, Artificial Intelligence Innovation Hub and No.2019-0-00075, Artificial Intelligence Graduate School Program(KAIST)), and the Engineering Research Center Program through the National Research Foundation of Korea (NRF) funded by the Korean Government MSIT (NRF-2018R1A5A1059921). We thank Geon Park for providing the visualization of diffusion in Figure 1, and Jinheon Baek for the suggestions on the experiments. 

%% file: 7_appendix.tex
\newpage
\appendix
\onecolumn
\begin{center}{\bf {\LARGE Appendix}}\end{center}
\vspace{0.15in}

\paragraph{Organization} The appendix is organized as follows: We first present the derivations excluded from the main paper due to space limitation in Section~\ref{sec:app:derivation}, and explain the details of the proposed score-based graph generation framework in Section~\ref{sec:app:generation_framework}. Then we provide the experimental details including the hyperparameters of the toy experiment, the generic graph generation, and the molecule generation in Section~\ref{sec:app:experiments}. Finally, we present additional experimental results and the visualizations of the generated graphs in Section~\ref{sec:app:additional}.

\section{Derivations \label{sec:app:derivation}}
In this section, we present the detailed derivations of the proposed training objectives described in Section~\ref{subsec:objectives} and the derivations of our novel S4 solver explained in Section~\ref{subsec:solve}.   

\subsection{Deriving the Denoising Score Matching Objectives}
\label{sec:app/matching}
The original score matching objective can be written as follows:
\begin{align}
    \mathbb{E}_{\bm{G}_t}\Big\|\bm{s}_{\theta}(\bm{G}_t,t)-\nabla_{\!\bm{X}_t}\log p_t(\bm{G}_t) \Big\|^2_2 = \mathbb{E}_{\bm{G}_t}\Big\|\bm{s}_{\theta}(\bm{G}_t,t) \Big\|^2_2 - 2\mathbb{E}_{\bm{G}_t}\Big\langle \bm{s}_{\theta}, \nabla_{\!\bm{X}_t}\log p_t(\bm{G}_t) \Big\rangle + C_1,
\end{align}
where $C_1$ is a constant that does not depend on $\theta$. Further, the denoising score matching objective can be written as follows:
\begin{equation}
    \begin{aligned}
    \mathbb{E}_{\bm{G}_0}\mathbb{E}_{\bm{G}_t|\bm{G}_0} \Big\|\bm{s}_{\theta}(\bm{G}_t,t)-\nabla_{\!\bm{X}_t}\log p_t(\bm{G}_t|\bm{G}_0) \Big\|^2_2 &= \mathbb{E}_{\bm{G}_0}\mathbb{E}_{\bm{G}_t|\bm{G}_0} \Big\|\bm{s}_{\theta}(\bm{G}_t,t) \Big\|^2_2 \\
    &- 2\mathbb{E}_{\bm{G}_0}\mathbb{E}_{\bm{G}_t|\bm{G}_0}\Big\langle \bm{s}_{\theta}, \nabla_{\!\bm{X}_t}\log p_t(\bm{G}_t|\bm{G}_0) \Big\rangle + C_2,
    \end{aligned}
\end{equation}
where $C_2$ is also a constant that does not depend on $\theta$.
From the following equivalence,
\begin{align*}
    \mathbb{E}_{\bm{G}_t}\Big\langle \bm{s}_{\theta}, \nabla_{\!\bm{X}_t}\log p_t(\bm{G}_t) \Big\rangle 
    &= \int_{\bm{G}_t}p(\bm{G}_t)\Big\langle\bm{s}_{\theta}, \nabla_{\!\bm{X}_t}\log p_t(\bm{G}_t) \Big\rangle\mathrm{d}\bm{G}_t \\
    &= \int_{\bm{G}_t}\Big\langle\bm{s}_{\theta}, \nabla_{\!\bm{X}_t}p_t(\bm{G}_t) \Big\rangle\mathrm{d}\bm{G}_t \\
    &= \int_{\bm{G}_t}\Big\langle\bm{s}_{\theta}, \nabla_{\!\bm{X}_t}\int_{\bm{G}_0}p(\bm{G}_0)p_t(\bm{G}_t|\bm{G}_0)\mathrm{d}\bm{G}_0 \Big\rangle\mathrm{d}\bm{G}_t \\
    &=\int_{\bm{G}_t}\Big\langle\bm{s}_{\theta}, \nabla_{\!\bm{X}_t}\int_{\bm{G}_0}p(\bm{G}_0)p_t(\bm{G}_t|\bm{G}_0)\mathrm{d}\bm{G}_0 \Big\rangle\mathrm{d}\bm{G}_t \\
    =& \int_{\bm{G}_t}\Big\langle\bm{s}_{\theta}, \int_{\bm{G}_0}p(\bm{G}_0)\nabla_{\!\bm{X}_t}p_t(\bm{G}_t|\bm{G}_0)\mathrm{d}\bm{G}_0 \Big\rangle\mathrm{d}\bm{G}_t \\
    =& \int_{\bm{G}_t}\int_{\bm{G}_0}p(\bm{G}_0)p(\bm{G}_t|\bm{G}_0)\Big\langle\bm{s}_{\theta},\nabla_{\!\bm{X}_t}\log p_t(\bm{G}_t|\bm{G}_0)\Big\rangle\mathrm{d}\bm{G}_0\mathrm{d}\bm{G}_t \\
    =& \mathbb{E}_{\bm{G}_0}\mathbb{E}_{\bm{G}_t|\bm{G}_0}\Big\langle\bm{s}_{\theta},\nabla_{\!\bm{X}_t}\log p_t(\bm{G}_t|\bm{G}_0)\Big\rangle,
\end{align*}
we can conclude that the two objectives are equivalent with respect to $\theta$:
\begin{align}
    \mathbb{E}_{\bm{G}_0}\mathbb{E}_{\bm{G}_t|\bm{G}_0} \Big\|\bm{s}_{\theta}(\bm{G}_t,t)-\nabla_{\!\bm{X}_t}\log p_t(\bm{G}_t|\bm{G}_0) \Big\|^2_2 = 
    \mathbb{E}_{\bm{G}_t}\Big\|\bm{s}_{\theta}(\bm{G}_t,t)-\nabla_{\!\bm{X}_t}\log p_t(\bm{G}_t) \Big\|^2_2 + C_2 - C_1.
\end{align}
Similarly, computing the gradient with respect to $\bm{A}_t$, we can show that the following two objectives are also equivalent with respect to $\phi$:
\begin{align}
    \mathbb{E}_{\bm{G}_0}\mathbb{E}_{\bm{G}_t|\bm{G}_0} \Big\|\bm{s}_{\phi}(\bm{G}_t,t)-\nabla_{\!\bm{A}_t}\log p_t(\bm{G}_t|\bm{G}_0) \Big\|^2_2 = 
    \mathbb{E}_{\bm{G}_t}\Big\|\bm{s}_{\phi}(\bm{G}_t,t)-\nabla_{\!\bm{A}_t}\log p_t(\bm{G}_t) \Big\|^2_2 + C_4 - C_3,
\end{align}
where $C_3$ and $C_4$ are constants that does not depend on $\phi$.

\subsection{Deriving New Objectives for GDSS \label{sec:app/kernel}}
It is enough to show that $\dscore{X}{G}$ is equal to $\dscore{X}{X}$. 
Using the chain rule, we can derive that $\nabla_{\!\bm{X}_t}\log p_{0t}(\bm{A}_t|\bm{A}_0)=0$:
\begin{align}
    \frac{\partial\log p_{0t}(\bm{A}_t|\bm{A}_0)}{\partial(\bm{X}_t)_{ij}} 
    &= \text{Tr}\bigg[ \nabla_{\!\bm{A}_t}\log p_{0t}(\bm{A}_t|\bm{A}_0) \underbrace{\frac{\partial\bm{A}_t}{\partial(\bm{X}_t)_{ij}}}_{=0} \bigg] = 0,
\end{align}
Therefore, we can conclude that $\dscore{X}{G}$ is equal to $\dscore{X}{X}$:
\begin{align}
    \nabla_{\!\bm{X}_t}\log p_{0t}(\bm{G}_t|\bm{G}_0) 
    = \nabla_{\!\bm{X}_t}\log p_{0t}(\bm{X}_t|\bm{X}_0) + 
    \underbrace{\nabla_{\!\bm{X}_t}\log p_{0t}(\bm{A}_t|\bm{A}_0)}_{=0} 
    = \dscore{X}{X}.
\end{align}
Similarly, computing the gradient with respect to $\bm{A}_t$, we can also show that $\nabla_{\!\bm{A}_t}\log p_{0t}(\bm{G}_t|\bm{G}_0)$ is equal to $\nabla_{\!\bm{A}_t}\log p_{0t}(\bm{A}_t|\bm{A}_0)$.

\subsection{The Action of the Fokker-Planck Operators and the Classical Propagator \label{sec:app:solve}}

\paragraph{Fokker-Planck Operators}
Recall the system of reverse-time SDEs of Eq.~\eqref{eq:estimated_system_sdes}:
\begin{equation}
\fontsize{9pt}{9pt}\selectfont
\begin{aligned}
    \begin{cases}
        \mathrm{d}\bm{X}_t \!= \eqmakebox[F]{$\mathbf{f}_{1,t}(\bm{X}_t) \mathrm{d}\overbar{t} + g_{1,t}\mathrm{d}\mathbf{\bar{w}}_1$} \; \eqmakebox[S]{$- g^2_{1,t}\bm{s}_{\theta,t}(\bm{X}_t,\!\bm{A}_t)\mathrm{d}\overbar{t}$} \\
        \mathrm{d}\bm{A}_t\, \!= \eqmakebox[LHS]{$\mathbf{f}_{2,t}(\bm{A}_t) \mathrm{d}\overbar{t} + g_{2,t}\mathrm{d}\mathbf{\bar{w}}_2$} \; \eqmakebox[S]{$- g^2_{2,t}\bm{s}_{\!\phi,t}(\bm{X}_t,\!\bm{A}_t)\mathrm{d}\overbar{t}$}
    \end{cases} \\[-1.2\normalbaselineskip]
    \underbrace{\eqmakebox[F]{\mathstrut}}_{F}
    \phantom{}
    \underbrace{\eqmakebox[S]{\mathstrut}}_{S}
    \phantom{a\;}
    \label{eq:estimated_system_sdes_appendix}
\end{aligned},
\fontsize{10pt}{10pt}\selectfont
\end{equation}
Denoting the marginal joint distribution of Eq.~\eqref{eq:estimated_system_sdes_appendix} at time $t$ as $\tilde{p}_t(\bm{G}_t)$, the evolution of $\tilde{p}_t$ through time $t$ can be described by a partial differential equation, namely Fokker-Planck equation, as follows:
\begin{align}
    \frac{\partial\tilde{p}_t(\bm{G}_t)}{\partial t} = -\nabla_{\!\bm{G}_t}\cdot\left( \mathbf{f}_{t}(\bm{G}_t)\tilde{p}_t(\bm{G}_t) - \frac{1}{2}g_t^2\tilde{p}_t(\bm{G}_t)\nabla_{\!\bm{G}_t}\log\tilde{p}_t(\bm{G}_t) - g_t^2\bm{s}_t(\bm{G}_t)\tilde{p}_t(\bm{G}_t) \right),
\end{align}
where $\bm{s}_t(\bm{G}_t) = (\bm{s}_{\theta,\!t}(\bm{G}_t),\!\bm{s}_{\phi,t}(\bm{G}_t))$.
Then, the Fokker-Planck equation can be represented using the Fokker-Planck operators as follows:
\begin{align}
    \frac{\partial\tilde{p}_t(\bm{G}_t)}{\partial t} = (\fop_{F} + \fop_{S})\tilde{p}_t(\bm{G}_t),
    \label{eq:fokker-planck}
\end{align}
where the action of the Fokker-Planck operators on the function $\bm{A}(\bm{G}_t)$ is defined as:
\begin{align}
    \fop_{F}\bm{A}(\bm{G}_t) &\coloneqq -\nabla_{\!\bm{G}_t}\cdot\left( \mathbf{f}_{t}(\bm{G}_t) \bm{A}(\bm{G}_t) - \frac{1}{2}g_t^2\bm{A}(\bm{G}_t)\nabla_{\!\bm{G}_t}\log\bm{A}(\bm{G}_t) \right) \\
    \fop_{S}\bm{A}(\bm{G}_t) &\coloneqq -\nabla_{\!\bm{G}_t}\cdot\left( - g_t^2\bm{s}_t(\bm{G}_t)\bm{A}(\bm{G}_t) \right).
    \label{eq:fokker-planck-operators}
\end{align}

\paragraph{Classical Propagator} 
From the Fokker-Planck equation of Eq.~\eqref{eq:fokker-planck}, we can derive an intractable solution $\bar{\bm{G}}_t\coloneqq\bm{G}_{T-t}$ to the system of reverse-time SDEs in Eq.~\eqref{eq:estimated_system_sdes_appendix} as follows:
\begin{align}
    \bar{\bm{G}}_t = e^{t(\fop_F+\fop_S)}\bar{\bm{G}}_{0},
\end{align}
which is called the classical propagator. The action of the operator $e^{t(\fop_F+\fop_S)}$ propagates the initial states $\bar{\bm{G}}_0$ to time $t$ following the dynamics determined by the action of Fokker-Planck operators $\fop_F$ and $\fop_S$, described in Eq.~\eqref{eq:fokker-planck-operators}.

\subsection{Symmetric Splitting for the System of SDEs \label{sec:app:s4_derivation}}
Let us take a look on each operator one by one. First, the action of the operator $\fop_{F}$ on the marginal distribution $\tilde{p}_t(\bm{G}_t)$ corresponds to the diffusion process described by the $F$-term in Eq.~\eqref{eq:estimated_system_sdes_appendix} as follows:
\begin{align}
    \begin{cases}
    \mathrm{d}\bm{X}_t = \mathbf{f}_{1,t}(\bm{X}_t)\mathrm{d}\overbar{t} + g_{1,t}\mathrm{d}\mathbf{\bar{w}_1} \\
    \mathrm{d}\bm{A}_t = \mathbf{f}_{2,t}(\bm{A}_t)\mathrm{d}\overbar{t} + g_{2,t}\mathrm{d}\mathbf{\bar{w}_2} \\
    \end{cases},
    \label{eq:action_F}
\end{align}
Notice that the diffusion process described by Eq.~\eqref{eq:action_F} is similar to the forward diffusion process of Eq.~\eqref{eq:forward_diffusion}, with a slight difference that Eq.~\eqref{eq:action_F} is a system of  reverse-time SDEs. 
Therefore, the action of the operator $e^{\frac{\delta t}{2}\fop_{F}}$ can be represented by the transition distribution of the forward diffusion process $p_{st}(\cdot|\cdot)$ as follows:
\begin{align}
    e^{\frac{\delta t}{2}\fop_F}\bm{G} = \tilde{\bm{G}}\sim p_{t,t-\delta t/2}(\tilde{\bm{G}}|\bm{G}).
\end{align}
We provide the explicit form of the transition distribution in Section~\ref{sec:app:transition_kernel} of the Appendix.
Furthermore, the operator $\fop_{S}$ corresponds to the evolution of $\bm{X}_t$ and $\bm{A}_t$ described by the $S$-term in Eq.~\eqref{eq:estimated_system_sdes_appendix}, which is a system of reverse-time ODEs:
\begin{align}
    \begin{cases}
    \bm{X}_t = -g^2_{1,t}\bm{s}_{\theta,t}(\bm{X}_t,\bm{A}_t)\mathrm{d}\overbar{t} \\
    \bm{A}_t = -g^2_{2,t}\bm{s}_{\phi,t}(\bm{X}_t,\bm{A}_t)\mathrm{d}\overbar{t} \\
    \end{cases},
\end{align}
Hence, the action of the operator $e^{\delta t\fop_{S}}$ can be approximated with the simple Euler method for a positive time step $\delta t$:
\begin{equation}
    \begin{aligned}
    e^{\delta t\fop_{S}}\bm{X}_t\approx \bm{X}_t + g^2_{1,t}\bm{s}_{\theta,t}(\bm{X}_t,\bm{A}_t)\delta t \\
    e^{\delta t\fop_{S}}\bm{A}_t\approx \bm{A}_t + g^2_{2,t}\bm{s}_{\phi,t}(\bm{X}_t,\bm{A}_t)\delta t, \\
    \end{aligned}
\end{equation}
which we refer to this approximated action as $e^{\delta t\mathcal{L}^{Euler}_S}$. Using the symmtric Trotter theorem~\cite{Trotter}, we can approximate the intractable solution $e^{t(\fop_F+\fop_S)}$ as follows~\cite{CLD}:
\begin{align}
    e^{t(\fop_F+\fop_S)} 
    &\approx\left[ e^{\frac{\delta t}{2}\fop_F} e^{\delta t\fop_S} e^{\frac{\delta t}{2}\fop_F}\right]^M + O(M\delta t^3) \notag \\
    &= \left[ e^{\frac{\delta t}{2}\fop_F} e^{\delta t\mathcal{L}^{Euler}_S} e^{\frac{\delta t}{2}\fop_F}\right]^M + MO(\delta t^2) \label{eq:local_error} \\
    &= \left[ e^{\frac{\delta t}{2}\fop_F} e^{\delta t\mathcal{L}^{Euler}_S} e^{\frac{\delta t}{2}\fop_F}\right]^M + O(\delta t), \label{eq:global_error}
\end{align}
for a sufficiently large number of steps $M$ and a time step $\delta t=t/M$. 
Note that from Eq.~\eqref{eq:local_error} and Eq.~\eqref{eq:global_error}, we can see that the prediction step of S4 has local error $\mathcal{O}(\delta t^2)$ and global error $\mathcal{O}(\delta t)$.
From the action of the Fokker-Planck operators and the result of Eq.~\eqref{eq:global_error}, we can derive the prediction step of the S4 solver described in Section~\ref{subsec:solve}.
We further provide the pseudo-code for the proposed S4 solver in Algorithm~\ref{alg:symmetric_splitting}.

\vspace{-0.3in}
\input{table/algorithm}
\vspace{0.2in}

\subsection{Derivation of the transition distribution \label{sec:app:transition_kernel}}
We provide an explicit form of the transition distribution for two types of SDE, namely VPSDE and VESDE~\cite{score-based/2}. We consider the transition distribution from time $t$ to $t-\delta t$ for sufficiently small time step $\delta t$ with $\bm{x}_t$ given, and considering the input as discrete state corresponding to normal distribution with 0 variance.

\paragraph{VPSDE}
The process of the VPSDE is given by the following SDE:
\begin{align}
    \mathrm{d}\bm{x} = -\frac{1}{2}\beta_t\bm{x}\mathrm{d}t + \sqrt{\beta_t}\mathrm{d}\mathbf{w},
    \label{eq:vpsde}
\end{align}
where $\beta_t=\beta_{min} + t(\beta_{max}-\beta_{min})$ for the hyperparameters $\beta_{min}$ and $\beta_{max}$, and $t\in[0,1]$. Since Eq.~\eqref{eq:vpsde} has a linear drift coefficient, the transition distribution of the process is Gaussian, and the mean and covariance can be derived using the result of Eq.(5.50) and (5.51) of \citet{kernel_derivation} as follows:
\begin{align}
    p_{t,t-\delta t}(\bm{x}_{t-\delta t}|\bm{x}_t) = \mathcal{N}\left(\bm{x}_{t-\delta t}\;|\;\mu_t\bm{x}_t, \bm{\Sigma_t}\right),
\end{align}
where $\mu_t = e^{C_t}$ and $\bm{\Sigma}_t = \bm{I} - \bm{I}e^{-2C_t}$ for
\begin{align}
    C_t = \frac{1}{2}\int^{t}_{t-\delta t}\beta_s\mathrm{d}s = \frac{\delta t}{4}\Big( 2\beta_{min} + (2t +\delta t)(\beta_{max}-\beta_{min}) \Big).
\end{align}

\paragraph{VESDE} 
The process of the VESDE is given by the following SDE:
\begin{align}
    \mathrm{d}\bm{x} = \sigma_{min}\left(\frac{\sigma_{max}}{\sigma_{min}}\right)^{t} \sqrt{2\log\frac{\sigma_{max}}{\sigma_{min}}}\mathrm{d}\mathbf{w},
    \label{eq:vesde}
\end{align}
for the hyperparameters $\sigma_{min}$ and $\sigma_{max}$, and $t\in(0,1]$. 
Since Eq.~\eqref{eq:vesde} has a linear drift coefficient, the transition distribution of the process is Gaussian, and the mean and covariance can be derived using the result of Eq.(5.50) and (5.51) of \citet{kernel_derivation} as follows:
\begin{align}
    p_{t,t-\delta t}(\bm{x}_{t-\delta t}|\bm{x}_t) = \mathcal{N}\left(\bm{x}_{t-\delta t}\;|\;\bm{x}_t, \bm{\Sigma_t}\right),
\end{align}
where $\bm{\Sigma_t} = \Sigma_t\bm{I}$ is given as:
\begin{align}
    \Sigma_t = \sigma_{min}^2\left(\frac{\sigma_{max}}{\sigma_{min}}\right)^{2t} - \sigma_{min}^2\left(\frac{\sigma_{max}}{\sigma_{min}}\right)^{2t-2\delta t}.
\end{align}

\section{Details for Score-based Graph Generation \label{sec:app:generation_framework}}
In this section, we describe the architectures of our proposed score-based models, and further provide the details of the graph generation procedure through the reverse-time diffusion process. 

\begin{figure*}[t!]
    \centering
    \includegraphics[width=\textwidth]{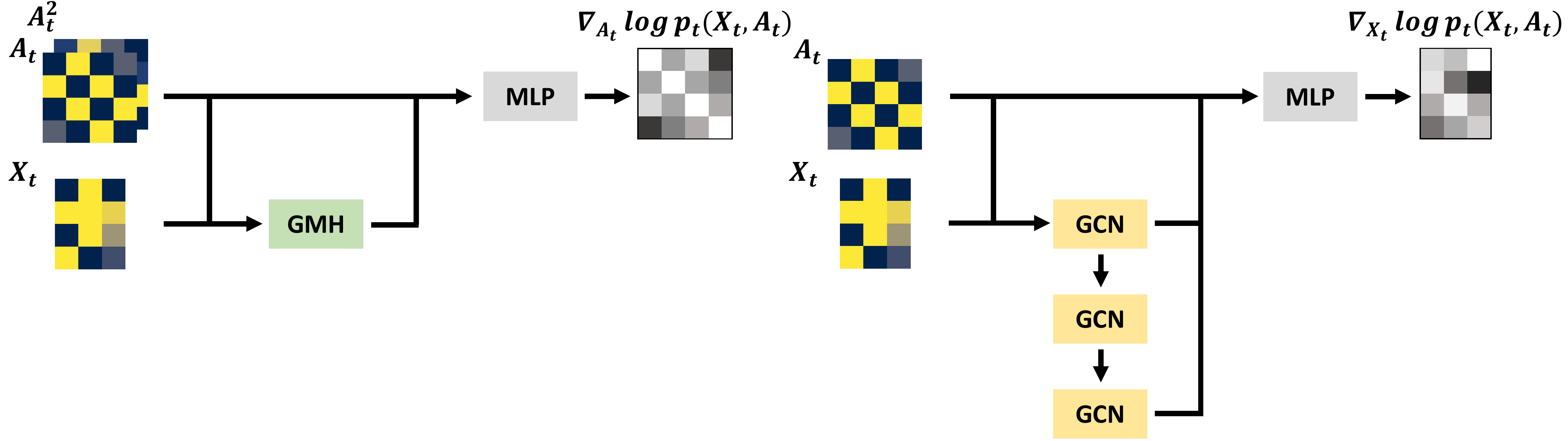}
    \caption{\small \textbf{The architecture of the score-based models of GDSS.} (Left) The score-based model $s_\phi$ estimating $\pscore{\bm{A}_t}{\bm{X}_t,\bm{A}_t}$ is composed of GMH blocks and MLP layers. (Right) The score-based model $s_\theta$ estimating $\pscore{\bm{X}_t}{\bm{X}_t,\bm{A}_t}$ is composed of GCN layers and MLP layers. Both models take $\bm{X_t}$ and $\bm{A_t}$ as input and estimate the partial scores with respect to $\bm{A_t}$ and $\bm{X_t}$, respectively.}
    \label{fig:model_architecture}
\end{figure*}

\subsection{Score-based Model Architecture}
We illustrate the architecture of the proposed score models $s_{\theta,t}$ and $s_{\phi,t}$ in Figure~\ref{fig:model_architecture}, which are described in Section~\ref{subsec:objectives:model}.

\subsection{Generating Samples from the Reverse Diffusion Process}
We first sample $N$, the number of nodes to be generated from the empirical distribution of the number of nodes in the training dataset as done in \citet{conditional} and \citet{score-based/graph/1}.
Then we sample the noise of batch size $B$ from the prior distribution, where $\bm{X}_{T}$ is of dimension $N\!\times\! F\!\times\! B$ and $\bm{A}_T$ is of dimension $N\!\times\! N\!\times\! B$, and simulate the reverse-time system of SDEs in Eq.~\eqref{eq:estimated_system_sdes} to obtain the solution $\bm{X}_0$ and $\bm{A}_0$. Lastly, we quantize $\bm{X}_0$ and $\bm{A}_0$ with the operation depending on the generation tasks. We provide further details of the generation procedure in Section~\ref{sec:app:experiments}, including the hyperparameters.

\section{Experimental Details \label{sec:app:experiments}}
In this section, we explain the details of the experiments including the toy experiments shown in Figure~\ref{fig:toy}, the generic graph generation tasks, and the molecule generation tasks. We describe the implementation details of GDSS and the baselines, and further provide the hyperparameters used in the experiments in Table~\ref{tab:params}.

\subsection{Toy Experiment \label{sec:app:toy}}
Here, we provide the details for the toy experiment presented in Section~\ref{subsec:diffusion}. We construct the distribution of the data with bivariate Gaussian mixture with the mean and the covariance as follows:
\begin{align}
    p_{data}(\mathbf{x}) &= \mathcal{N}(\mathbf{x}\;|\;\mu_1,\Sigma_1) + \mathcal{N}(\mathbf{x}\;|\;\mu_2,\Sigma_2), \\
    \mu_1 &= \begin{pmatrix}
    0.5 \\
    0.5 \\
    \end{pmatrix}
    \;,\; \mu_2 = \begin{pmatrix}
    -0.5 \\
    -0.5 \\
    \end{pmatrix}\;,\;
    \Sigma_1 = \Sigma_2 = 0.1^2\begin{pmatrix}
    1.0 & 0.9 \\
    0.9 & 1.0 \\
    \end{pmatrix}. \notag
\end{align}
For each diffusion method, namely GDSS, GDSS-seq, and independent diffusion, we train two models, where each model estimates the partial score or score with respect to the variable. We fix the number of linear layers in the model to 20 with the residual paths, and set the hidden dimension as 512. We use VPSDE for the diffusion process of each variable with $\beta_{min}=0.01$ and $\beta_{max}=0.05$. We train the models for 5000 epochs with batch size 2048 sampled from the data distribution. We generate $2^{13}$ samples for each diffusion method, shown in Figure~\ref{fig:toy}.

\input{table/hyperparameters}

\subsection{Generic Graph Generation \label{app:sec:generic_graph}}
The information and the statistics of the graph datasets, namely Ego-small, Community-small, Enzymes and Grid, are shown in Section~\ref{exp:generic_graph} and Table~\ref{tab:mmd}. We carefully selected the datasets to have varying sizes and characteristics, for example synthetic graphs, real-world graphs, social graphs or biochemical graphs. 

\vspace{-0.05in}
\paragraph{Implementation Details}
For a fair evaluation of the generic graph generation task, we follow the standard setting of existing works~\cite{you2018graphrnn, GNF, score-based/graph/1} from the node features to the data splitting.
Especially, for Ego-small and Community-small datasets, we report the means of 15 runs, 3 different runs for 5 independently trained models. For Enzymes and Grid dataset, since the baselines including GraphVAE and EDP-GNN take more than 3 days for a single training, we report the means of 3 different runs. For the baselines, we use the hyperparameters given by the original work, and further search for the best performance if none exists.
For GDSS, we initialize the node features as the one-hot encoding of the degrees. We perform the grid search to choose the best signal-to-noise ratio (SNR) in $\{0.05, 0.1, 0.15, 0.2\}$ and the scale coefficient in the $\{0.1, 0.2, 0.3, 0.4, 0.5, 0.6, 0.7, 0.8, 0.9, 1.0\}$. We select the best MMD with the lowest average of three graph statistics, degree, clustering coefficient, and orbit. Further, we observed that applying the exponential moving average (EMA)~\citet{score-based/1} for larger graph datasets, namely Enzymes and Grid, improves the performance and lowers the variance. After generating the samples by simulating the reverse diffusion process, we quantize the entries of the adjacency matrices with the operator $1_{x>0.5}$ to obtain the 0-1 adjacency matrix. We empirically found that the entries of the resulting samples after the simulation of the diffusion process do not deviate much from the integer values 0 and 1. We report the hyperparameters used in the experiment in Table~\ref{tab:params}.

\input{table/datasets}
\subsection{Molecule Generation}
The statistics of the molecular datasets, namely, QM9 and ZINC250k datasets, are summarized in Table~\ref{tab:data}.

\vspace{-0.05in}
\paragraph{Implementation Details of GDSS and GDSS-seq} \label{sec:app/mol/details}
Each molecule is preprocessed into a graph with the node features $X \!\in\! \{0, 1\}^{N \!\times\! F}$ and the adjacency matrix $A \!\in\! \{0, 1, 2, 3\}^{N \!\times\! N}$, where $N$ is the maximum number of atoms in a molecule of the dataset, and $F$ is the number of possible atom types. The entries of $A$ indicate the bond types, i.e. single, double, or triple bonds. Following the standard procedure~\cite{shi2020graphaf,luo2021graphdf}, the molecules are kekulized by the RDKit library~\cite{landrum2016rdkit} and hydrogen atoms are removed. As explained in Section~\ref{molecule_generation}, we make use of the valency correction proposed by \citet{zang2020moflow}.
We perform the grid search to choose the best signal-to-noise ratio (SNR) in $\{0.1, 0.2\}$ and the scale coefficient in $\{0.1, 0.2, 0.3, 0.4, 0.5, 0.6, 0.7, 0.8, 0.9, 1.0\}$. Since the low novelty value leads to low FCD and NSPDK MMD values even though it is undesirable and meaningless, we choose the hyperparameters that exhibit the best FCD value among those which show the novelty that exceeds 85\%. After generating the samples by simulating the reverse diffusion process, we quantize the entries of the adjacency matrices to $\{0,1,2,3\}$ by clipping the values as: $(-\infty, 0.5)$ to 0, the values of $[0.5, 1.5)$ to 1, the values of $[1.5, 2.5)$ to 2, and the values of $[2.5, +\infty)$ to 3. We empirically observed that the entries of the resulting samples after the simulation of the diffusion process do not deviate much from the integer values 0, 1, 2, and 3. We report the hyperparameters used in the experiment in Table~\ref{tab:params}.

\vspace{-0.05in}
\paragraph{Implementation Details of Baselines}
We utilize the DIG~\cite{dig} library to generate molecules with GraphDF and GraphEBM. To conduct the experiments with GraphAF, we use the DIG library for the QM9 dataset, and use the official code\footnote{https://github.com/DeepGraphLearning/GraphAF} for the ZINC250k dataset. We use the official code\footnote{https://github.com/calvin-zcx/moflow} for MoFlow. We follow the experimental settings reported in the respective original papers and the codes for these models. For EDP-GNN, we use the same preprocessing and postprocessing procedures as in GDSS and GDSS-seq, except that we divide the adjacency matrices by 3 to ensure the entries are in the range of $[0, 1]$ before feeding them into the model. We conduct the grid search to choose the best size of the Langevin step in $\{0.01, 0.005, 0.001\}$ and noise scale in $\{0.1, 0.2, 0.3, 0.4, 0.5, 0.6, 0.7, 0.8, 0.9, 1.0\}$, and apply the same tuning criterion as in GDSS and GDSS-seq.

\subsection{Computing Resources}
For all the experiments, we utilize PyTorch~\cite{pytorch} to implement GDSS and train the score models on TITAN XP, TITAN RTX, GeForce RTX 2080 Ti, and GeForce RTX 3090 GPU.
For the generic graph generation tasks, the time comparison between the SDE solvers in Figure~\ref{fig:vis} was measured on 1 GeForce RTX 2080 Ti GPU and 40 CPU cores. For the molecule generation tasks, the inference time of each model is measured on 1 TITAN RTX GPU and 20 CPU cores.

\section{Additional Experimental Results \label{sec:app:additional}}
In this section, we provide additional experimental results.

\input{table/mmd_variance}
\input{table/mmd_variance_2}

\subsection{Generic Graph Generation \label{sec:app:exp_generic_graph}}
We report the standard deviation of the generation results of Table~\ref{tab:mmd} in Table~\ref{tab:mmd_variance_1} and Table~\ref{tab:mmd_variance_2}.

\setlength{\columnsep}{5pt}%
\begin{wraptable}{h}{0.6\textwidth}
\vspace{-0.15in}
\centering
\caption{\small \textbf{Generation results of MMD using a larger number (1024) of samples.}}
\vspace{-0.1in}
\input{table/mmd_1024}
\label{tab:1024}
\vspace{-0.1in}
\end{wraptable}
For a fair evaluation of the generative methods, following \citet{you2018graphrnn}, we have measured MMD between the test datasets and the set of generated graphs that have the same number of graphs as the test datasets. To further compare extensively with the baselines, following \citet{GNF, score-based/graph/1}, we provide the results of MMD measured between the test datasets and the set of 1024 generated graphs in Table~\ref{tab:1024}. We can observe that GDSS still outperforms the baselines using a larger number of samples (1024) to measure the MMD, and significantly outperforms EDP-GNN.

\subsection{Molecule Generation \label{sec:app:exp_mol}}
We additionally report the validity, uniqueness, and novelty of the generated molecules as well as the standard deviation of the results in Table~\ref{tab:qm9} and Table~\ref{tab:zinc250k}. \textbf{Validity} is the fraction of the generated molecules that do not violate the chemical valency rule. \textbf{Uniqueness} is the fraction of the valid molecules that are unique. \textbf{Novelty} is the fraction of the valid molecules that are not included in the training set.

\input{table/qm9}
\input{table/zinc250k}

\subsection{Ablation Studies \label{sec:app:ablation}}

\input{table/solver_appendix}

In Table~\ref{tab:S4_full}, we provide the full results of the table in Figure~\ref{fig:vis}, which shows the comparison between the fixed step size SDE solvers on other datasets. As shown in Table~\ref{tab:S4_full}, S4 significantly outperforms the predictor-only methods, and further shows competitive results compared to the PC samplers with half the computation time.

\section{Visualization}
In this section, we additionally provide the visualizations of the generated graphs for the generic graph generation tasks and molecule generation tasks.

\subsection{Generic Graph Generation \label{sec/app/vis_generic}}
We visualize the graphs from the training datasets and the generated graphs of GDSS for each datasets in Figure 6-9. The visualized graphs are the randomly selected samples from the training datasets and the generated graph set. We additionally provide the information of the number of edges $e$ and the number of nodes $n$ of each graph. 

\begin{figure}[ht]
    \centering
    \includegraphics[width=0.95\linewidth]{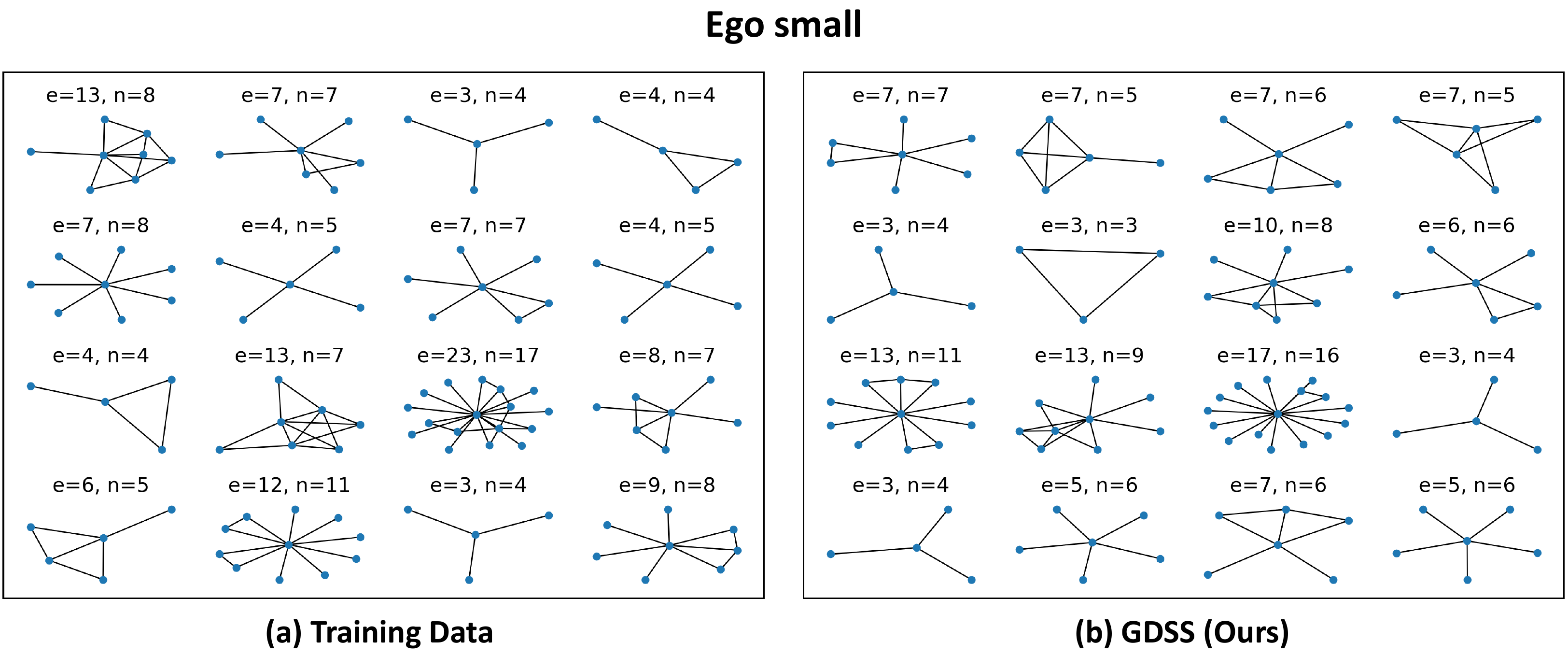}
    \vspace{-0.1in}
    \caption{\small \textbf{Visualization of the graphs from the Ego small dataset and the generated graphs of GDSS.}}
    \label{fig:ego_small}
\end{figure}

\begin{figure}[ht]
    \centering
    \includegraphics[width=0.95\linewidth]{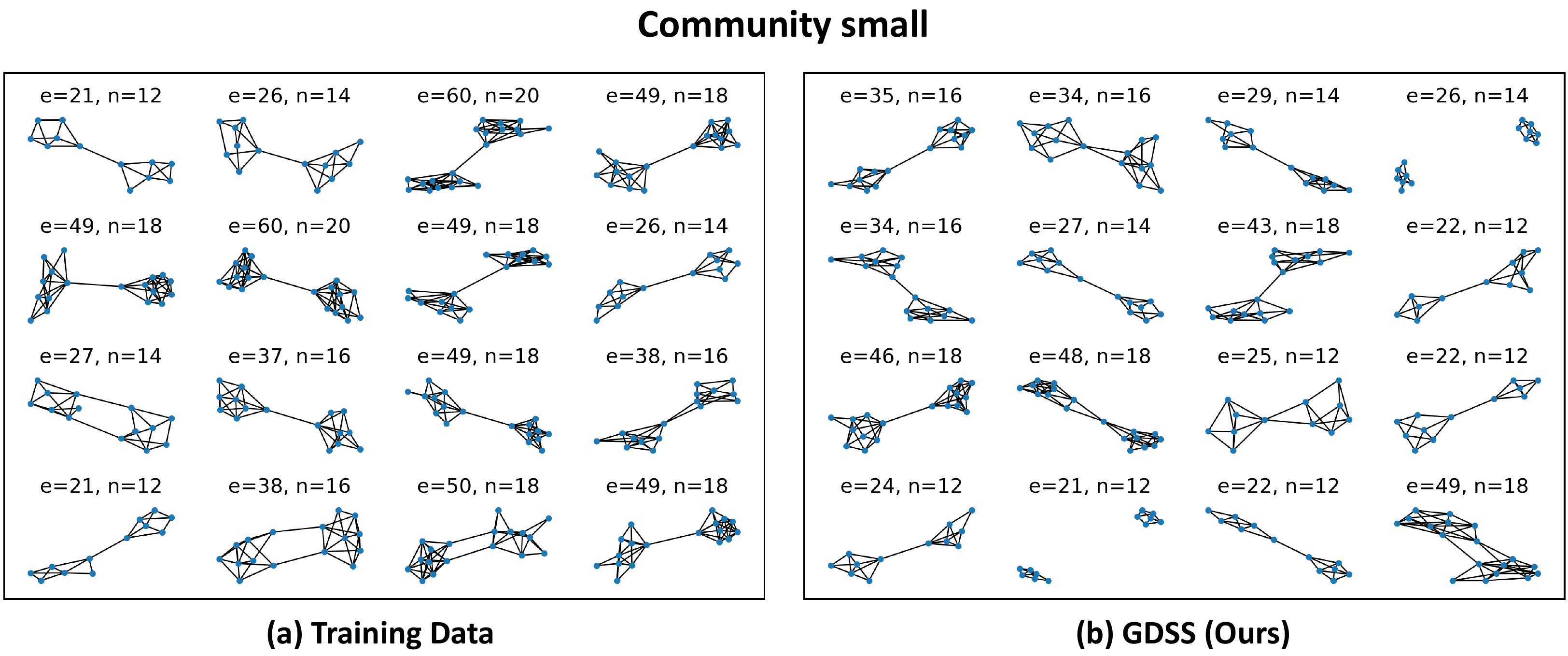}
    \vspace{-0.1in}
    \caption{\small \textbf{Visualization of the graphs from the Community small dataset and the generated graphs of GDSS.}}
    \label{fig:community_small}
\end{figure}

\begin{figure}[ht]
    \centering
    \includegraphics[width=0.95\linewidth]{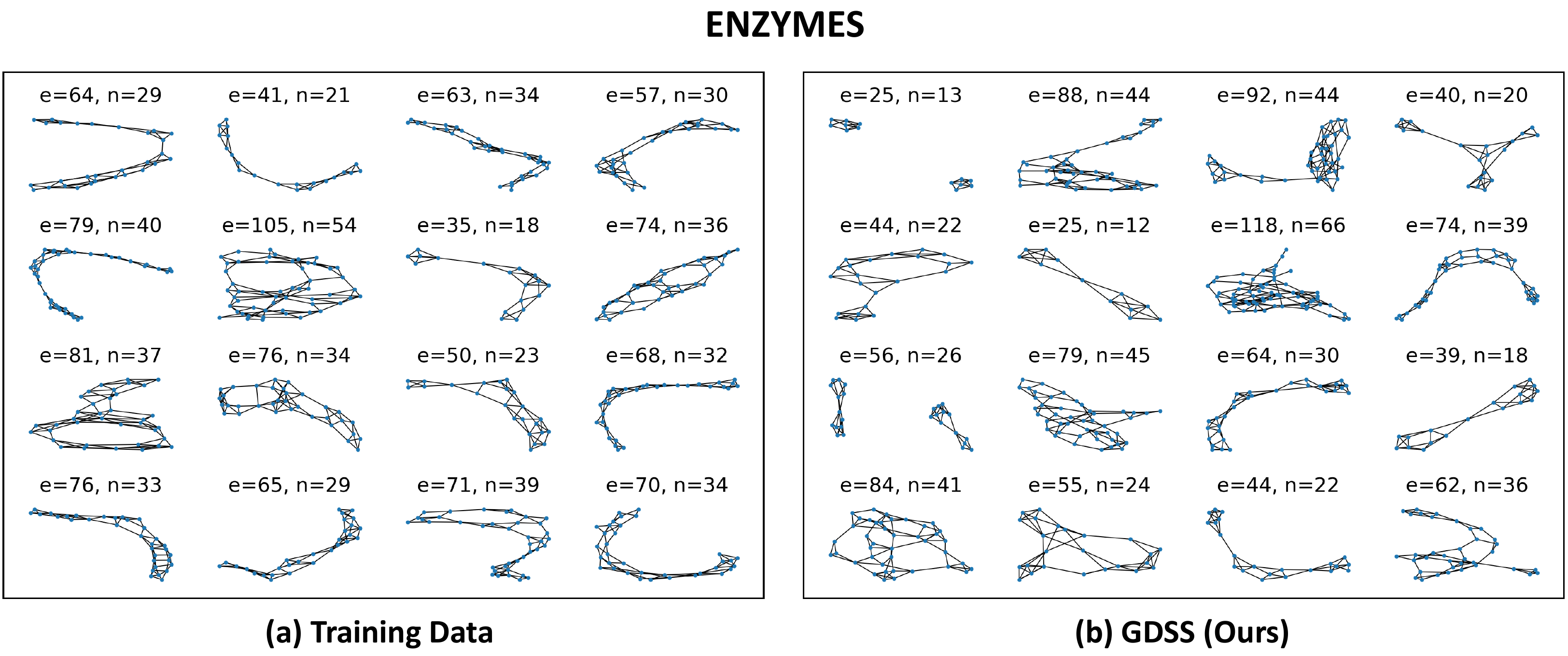}
    \vspace{-0.1in}
    \caption{\small \textbf{Visualization of the graphs from the ENZYMES dataset and the generated graphs of GDSS.}}
    \label{fig:enzymes}
\end{figure}

\clearpage

\begin{figure}[ht]
    \centering
    \includegraphics[width=0.95\linewidth]{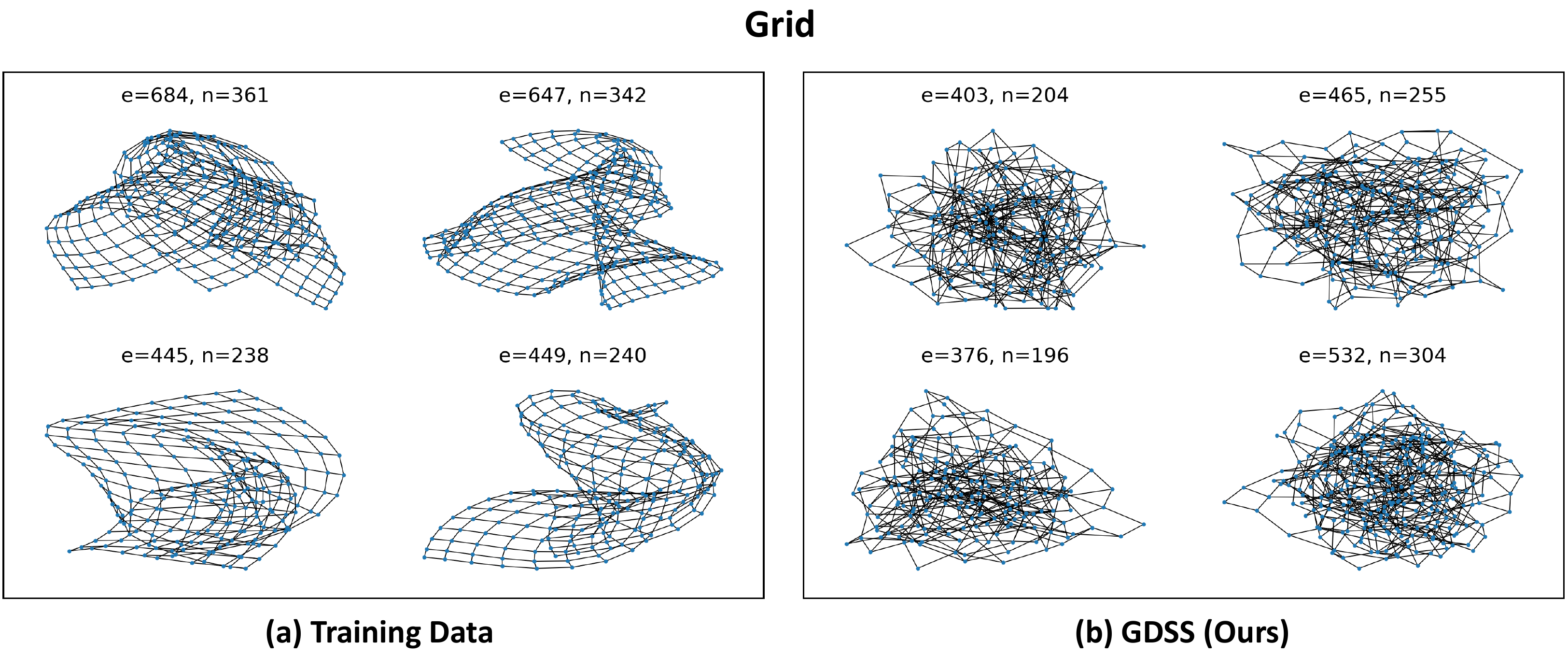}
    \vspace{-0.1in}
    \caption{\small \textbf{Visualization of the graphs from the Grid dataset and the generated graphs of GDSS.}}
    \label{fig:grid}
\end{figure}

\subsection{Molecule Generation \label{sec/app/vis_mol}}
\label{sec:app/vis_mol}
\begin{figure}[ht]
    \centering
    \includegraphics[width=0.98\linewidth]{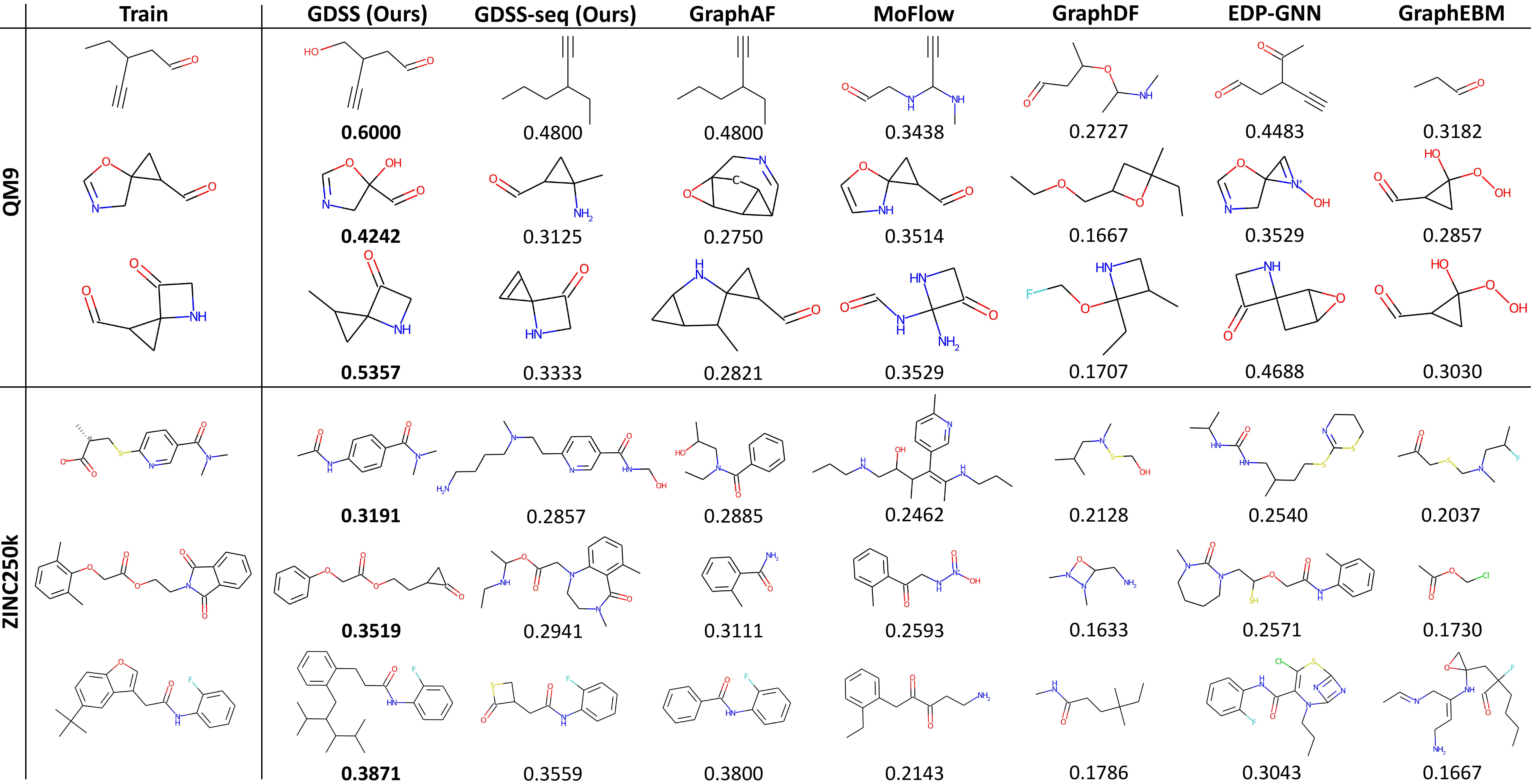}
    \caption{\small \textbf{Visualization of generated molecules with maximum Tanimoto similarity} with the molecule from the dataset. For each generated molecule, we display the similarity value at the bottom. }
    \label{fig:mols_app}
\end{figure}

We visualize the generated molecules that are maximally similar to certain training molecules in Figure~\ref{fig:mols_app}. The similarity measure is the Tanimoto similarity based on the Morgan fingerprints, which are obtained by the RDKit~\cite{landrum2016rdkit} library with radius 2 and 1024 bits. As shown in the figure, GDSS is able to generate molecules that are structurally close to the training molecules while other baselines generate molecules that deviate from the training distribution.

%% file: table/algorithm.tex
\begin{figure}[t!]
\vspace{-0.2in}
\centering
\begin{minipage}{0.9\linewidth}
\centering
\begin{algorithm}[H]
    \small
    \caption{ Symmetric Splitting for the System of SDEs (S4) }\label{alg:symmetric_splitting}
        \textbf{Input:} Score-based models $\bm{s}_{\theta,t}$ and $\bm{s}_{\phi,t}$, number of sampling steps $M$, step size $\delta t$, transition distributions $p_{st}(\cdot|\cdot)$ of the forward diffusion in Eq.~\eqref{eq:forward_diffusion}, Lagevin MCMC step size $\alpha$, scaling coefficient $\epsilon_s$ \\
        \textbf{Output:} $\bm{X}_0$, $\bm{A}_0$: the solution to Eq.~\eqref{eq:estimated_system_sdes}
    \begin{algorithmic}[1]
        \STATE $t=T$
        \STATE Sample from the prior distribution $\bm{X}_M,\!\bm{A}_M\sim p_T$
        \FOR{$m=M-1$ \textbf{to} $0$}
            \STATE $\bm{S}_{X} \leftarrow \bm{s}_{\theta,t}(\bm{X}_{m+1},\!\bm{A}_{m+1})$; \;\;
            $\bm{S}_A \leftarrow \bm{s}_{\phi,t}(\bm{X}_{m+1},\!\bm{A}_{m+1})$ \COMMENT{score computation step}
            \STATE $\bm{X}_{m+1}\leftarrow \bm{X}_{m+1} + \frac{\alpha}{2}\bm{S}_X + \epsilon_s \sqrt{\alpha}\bm{z}_X$; \;\; $\bm{z}_X\sim\mathcal{N}\left(\bm{0},\bm{I}_{X}\right)$ \COMMENT{correction step: $\bm{X}$}
            \STATE $\bm{A}_{m+1}\,\leftarrow \bm{A}_{m+1}\, + \frac{\alpha}{2}\bm{S}_A + \epsilon_s \sqrt{\alpha}\bm{z}_A$; \;\; $\bm{z}_A\sim\mathcal{N}\left(\bm{0},\bm{I}_{A}\right)$ \COMMENT{correction step: $\bm{A}$}
            \STATE $t' \leftarrow t - \delta t/2$
            \STATE $\tilde{\bm{X}}_{m}\sim p_{t,t'}(\tilde{\bm{X}}_{m}|\bm{X}_{m+1})$; \;\; $\tilde{\bm{A}}_{m}\sim p_{t,t'}(\tilde{\bm{A}}_{m}|\bm{A}_{m+1})$ \COMMENT{prediction step: action of $e^{\frac{\delta t}{2}\fop_{F}}$}
            \STATE $\tilde{\bm{X}}_{m}\leftarrow\tilde{\bm{X}}_{m} + g^2_{1,t}\bm{S}_{X}\delta t$; \;\; $\tilde{\bm{A}}_{m}\leftarrow\tilde{\bm{A}}_{m} + g^2_{2,t}\bm{S}_{A}\delta t$ \COMMENT{prediction step: action of $e^{\delta t\fop_{S}}\,\,$}
            \STATE $t\leftarrow t-\delta t$
            \STATE $\bm{X}_{m}\sim p_{t',t}(\bm{X}_{m}|\tilde{\bm{X}}_{m})$; \, $\bm{A}_{m}\sim p_{t',t}(\bm{A}_{m}|\tilde{\bm{A}}_{m})$ \COMMENT{prediction step: action of $e^{\frac{\delta t}{2}\fop_{F}}$}
        \ENDFOR
        \STATE \textbf{Return:} $\bm{X}_0$, $\bm{A}_0$ 
    \end{algorithmic}
\end{algorithm}
\end{minipage}
\vspace{-0.2in}
\end{figure}

%% file: table/hyperparameters.tex
\begin{table*}[t!]
\vspace{-0.05in}
    \caption{\small \textbf{Hyperparameters of GDSS} used in the generic graph generation tasks and the molecule generation tasks. We provide the hyperparameters of the score-based models ($\bm{s}_{\theta}$ and $\bm{s}_{\phi}$), the diffusion processes (SDE for $\bm{X}$ and $\bm{A}$), the SDE solver, and the training.}
    \vspace{-0.1in}
    \centering
    \resizebox{\textwidth}{!}{
    \begin{tabular}{llcccccc}
    \toprule
        & Hyperparameter & Ego-small & Community-small & Enzymes & Grid & QM9 & ZINC250k \\
    \midrule
        \multirow{2}{*}{$\bm{s}_\theta$}
        & Number of GCN layers & 2 & 3 & 5 & 5 & 2 & 2 \\
        & Hidden dimension & 32 & 32 & 32 & 32 & 16 & 16 \\
    \midrule
        \multirow{6}{*}{$\bm{s}_\phi$}
        & Number of attention heads & 4 & 4 & 4 & 4 & 4 & 4 \\
        & Number of initial channels & 2 & 2 & 2 & 2 & 2 & 2 \\
        & Number of hidden channels & 8 & 8 & 8 & 8 & 8 & 8 \\
        & Number of final channels & 4 & 4 & 4 & 4 & 4 & 4 \\
        & Number of GCN layers & 5 & 5 & 7 & 7 & 3 & 6 \\
        & Hidden dimension & 32 & 32 & 32 & 32 & 16 & 16 \\
    \midrule
        \multirow{4}{*}{SDE for $\bm{X}$}
        & Type & VP & VP & VP & VP & VE & VP \\
        & Number of sampling steps & 1000 & 1000 & 1000 & 1000 & 1000 & 1000 \\
        & $\beta_{min}$ & 0.1 & 0.1 & 0.1 & 0.1 & 0.1 & 0.1 \\
        & $\beta_{max}$ & 1.0 & 1.0 & 1.0 & 1.0 & 1.0 & 1.0 \\
    \midrule
        \multirow{4}{*}{SDE for $\bm{A}$}
        & Type & VP & VP & VE & VP & VE & VE \\
        & Number of sampling steps & 1000 & 1000 & 1000 & 1000 & 1000 & 1000 \\
        & $\beta_{min}$ & 0.1 & 0.1 & 0.2 & 0.2 & 0.1 & 0.2 \\
        & $\beta_{max}$ & 1.0 & 1.0 & 1.0 & 0.8 & 1.0 & 1.0 \\
    \midrule
        \multirow{3}{*}{Solver}
        & Type & EM & EM + Langevin & S4 & Rev. + Langevin  & Rev. + Langevin & Rev. + Langevin \\
        & SNR & - & 0.05 & 0.15 & 0.1 & 0.2 & 0.2 \\
        & Scale coefficient & - & 0.7 & 0.7 & 0.7 & 0.7 & 0.9 \\
    \midrule
        \multirow{6}{*}{Train} 
        & Optimizer & Adam & Adam & Adam & Adam & Adam & Adam \\
        & Learning rate & $1 \times 10^{-2}$ & $1 \times 10^{-2}$ & $1 \times 10^{-2}$ & $1 \times 10^{-2}$ & $5 \times 10^{-3}$ & $5 \times 10^{-3}$ \\
        & Weight decay & $1 \times 10^{-4}$ & $1 \times 10^{-4}$ & $1 \times 10^{-4}$ & $1 \times 10^{-4}$ & $1 \times 10^{-4}$ & $1 \times 10^{-4}$ \\
        & Batch size & 128 & 128 & 64 & 8 & 1024 & 1024 \\
        & Number of epochs & 5000 & 5000 & 5000 & 5000 & 300 & 500 \\
        & EMA & - & - & 0.999 & 0.999 & - & - \\
    \bottomrule
    \end{tabular}}
    \label{tab:params}
\vspace{-0.1in}
\end{table*}

%% file: table/datasets.tex
\begin{table*}[t!]
    \caption{\small\textbf{Statistics of QM9 and ZINC25ok datasets} used in the molecule generation tasks.}
\vspace{-0.1in}
    \centering
    \resizebox{0.8\textwidth}{!}{
    \begin{tabular}{lcccc}
    \toprule
        Dataset & Number of graphs & Number of nodes & Number of node types & Number of edge types \\
    \midrule
        QM9 & 133,885 & 1 $\le \vert V \vert \le$ 9 & 4 & 3 \\
        ZINC250k & 249,455 & 6 $\le \vert V \vert \le$ 38 & 9 & 3 \\
    \bottomrule
    \end{tabular}}
    \label{tab:data}
\vspace{-0.1in}
\end{table*}

%% file: table/mmd_variance.tex
\begin{table*}[t!]
\vspace{-0.05in}
    \caption{\small \textbf{Generation results of GDSS on the Ego-small and the Community-small datasets.} We report the MMD distance between the test datasets and generated graphs with the standard deviation.}
    \label{tab:mmd_variance_1}
\vspace{-0.1in}
    \centering
    \resizebox{0.9\textwidth}{!}{
    \renewcommand{\arraystretch}{1.0}
    \renewcommand{\tabcolsep}{8pt}
    \begin{tabular}{lcccccc}
    \toprule
        & \multicolumn{3}{c}{{Ego-small}} & \multicolumn{3}{c}{{Community-small}} \\
    \cmidrule(l{2pt}r{2pt}){2-4}
    \cmidrule(l{2pt}r{2pt}){5-7}
        &
        \multicolumn{3}{c}{Real, $4\leq|V|\leq18$} &
        \multicolumn{3}{c}{Synthetic, $12\leq|V|\leq20$} \\
    \cmidrule(l{2pt}r{2pt}){2-4}
    \cmidrule(l{2pt}r{2pt}){5-7}
        & Deg. & Clus. & Orbit & Deg. & Clus. & Orbit \\
    \midrule
        \textbf{GDSS-seq} (Ours) & 0.032 $\pm$ 0.006 & 0.027 $\pm$ 0.005 & 0.011 $\pm$ 0.007 & 0.090 $\pm$ 0.021 & 0.123 $\pm$ 0.200 & \textbf{0.007} $\pm$ 0.003 \\
        \textbf{GDSS} (Ours) & \textbf{0.021} $\pm$ 0.008 & \textbf{0.024} $\pm$ 0.007 & 0.007 $\pm$ 0.005 & \textbf{0.045} $\pm$ 0.028 & \textbf{0.086} $\pm$ 0.022 & \textbf{0.007} $\pm$ 0.004 \\
    \bottomrule
    \end{tabular}}
\vspace{-0.1in}
\end{table*}

%% file: table/mmd_variance_2.tex
\begin{table*}[t!]
    \caption{\small \textbf{Generation results on the Enzymes and the Grid datasets.} We report the MMD distance between the test datasets and generated graphs with the standard deviation. Best results are highlighted in bold (smaller the better). Hyphen (-) denotes out-of-resources that take more than 10 days or not applicable due to memory issue. $^*$ denotes our own implementation.}
    \label{tab:mmd_variance_2}
\vspace{-0.1in}
    \centering
    \resizebox{0.9\textwidth}{!}{
    \renewcommand{\arraystretch}{1.0}
    \renewcommand{\tabcolsep}{8pt}
    \begin{tabular}{lcccccc}
    \toprule
        & \multicolumn{3}{c}{{Enzymes}} &
        \multicolumn{3}{c}{{Grid}} \\
    \cmidrule(l{2pt}r{2pt}){2-4}
    \cmidrule(l{2pt}r{2pt}){5-7}
        &
        \multicolumn{3}{c}{Real, $10\leq|V|\leq125$} &
        \multicolumn{3}{c}{Synthetic, $100\leq|V|\leq400$} \\
    \cmidrule(l{2pt}r{2pt}){2-4}
    \cmidrule(l{2pt}r{2pt}){5-7}
        & Deg. & Clus. & Orbit & Deg. & Clus. & Orbit \\
    \midrule
        GraphRNN & \textbf{0.017} $\pm$ 0.007 & 0.062 $\pm$ 0.020 & 0.046 $\pm$ 0.031 & \textbf{0.064} $\pm$ 0.017 & 0.043 $\pm$ 0.022 & \textbf{0.021} $\pm$ 0.007 \\
        GraphAF$^*$ & 1.669 $\pm$ 0.024 & 1.283 $\pm$ 0.019 & 0.266 $\pm$ 0.007 & - & - & -  \\
        GraphDF$^*$ & 1.503 $\pm$ 0.011 & 1.061 $\pm$ 0.011 & 0.202 $\pm$ 0.002 & - & - & - \\
    \midrule
        GraphVAE$^*$ & 1.369 $\pm$ 0.020 & 0.629 $\pm$ 0.005 & 0.191 $\pm$ 0.020 & 1.619 $\pm$ 0.007 & 0.0 $\pm$ 0.000 & 0.919 $\pm$ 0.002 \\
        EDP-GNN & 0.023 $\pm$ 0.012 & 0.268 $\pm$ 0.164 & 0.082 $\pm$ 0.078 & 0.455 $\pm$ 0.319 & 0.238 $\pm$ 0.380 & 0.328 $\pm$ 0.278  \\
    \midrule
        \textbf{GDSS-seq} (Ours) & 0.099 $\pm$ 0.083 & 0.225 $\pm$ 0.051 & 0.010 $\pm$ 0.007 & 0.171 $\pm$ 0.134 & 0.011 $\pm$ 0.001 & 0.223 $\pm$ 0.070 \\
        \textbf{GDSS} (Ours) & 0.026 $\pm$ 0.008 & \textbf{0.061} $\pm$ 0.010 & \textbf{0.009} $\pm$ 0.005 & 0.111 $\pm$ 0.012 & \textbf{0.005} $\pm$ 0.000 & 0.070 $\pm$ 0.044 \\
    \bottomrule
    \end{tabular}}
\vspace{-0.1in}
\end{table*}

%% file: table/mmd_1024.tex
\centering
\resizebox{0.6\textwidth}{!}{
\renewcommand{\arraystretch}{0.9}
\renewcommand{\tabcolsep}{8pt}
\begin{tabular}{l c c c c c c c c c c c c c c c}
\toprule
    & 
    \multicolumn{3}{c}{{Ego-small}} &
    \multicolumn{3}{c}{{Community-small}} &
    \multirow{2}{*}{Avg.} \\
\cmidrule(l{2pt}r{2pt}){2-4}
\cmidrule(l{2pt}r{2pt}){5-7}
    & Deg. & Clus. & Orbit & Deg. & Clus. & Orbit & \\
\midrule
    GraphRNN & 0.040 & 0.050 & 0.060 & 0.030 & \textbf{0.010} & 0.010 & 0.033 \\
    GNF & \textbf{0.010} & 0.030 & \textbf{0.001} & 0.120 & 0.150 & 0.020 & 0.055 \\
    EDP-GNN & \textbf{0.010} & 0.025 & 0.003 & 0.006 & 0.127 & 0.018 & 0.031 \\
\midrule
    \textbf{GDSS} (Ours) & 0.023 & \textbf{0.020} & 0.005 & \textbf{0.029} & 0.068 & \textbf{0.004} & \textbf{0.030}  \\
\bottomrule
\end{tabular}}

%% file: table/qm9.tex
\begin{table*}[t!]
    \caption{\small \textbf{Generation results on the QM9 dataset.} Results are the means and the standard deviations of 3 runs. Values denoted by * are taken from the respective original papers. Other results are obtained by running open-source codes. Best results are highlighted in bold.}
\vspace{-0.1in}
    \centering
    \resizebox{\textwidth}{!}{
    \begin{tabular}{llc@{\hskip 0.01in}ccccccc}
    \toprule
        & \multirow{2}{*}{Method} & Validity w/o & \multirow{2}{*}{$\uparrow$} & NSPDK \multirow{2}{*}{$\downarrow$} & \multirow{2}{*}{FCD $\downarrow$} & \multirow{2}{*}{Validity (\%) $\uparrow$} & \multirow{2}{*}{Uniqueness (\%) $\uparrow$} & \multirow{2}{*}{Novelty (\%) $\uparrow$} \\
        & & correction (\%) & & MMD & & & & \\
    \midrule
        \multirow{4}{*}{Autoreg.}
        & GraphAF~\cite{shi2020graphaf} & 67* & & 0.020$\pm$0.003 & 5.268$\pm$0.403 & \textbf{100.00}* & 94.51* & 88.83* \\
        & GraphAF+FC & 74.43$\pm$2.55 & & 0.021$\pm$0.003 & 5.625$\pm$0.259 & \textbf{100.00}$\pm$0.00 & 88.64$\pm$2.37 & 86.59$\pm$1.95 \\
        & GraphDF~\cite{luo2021graphdf} & 82.67* & & 0.063$\pm$0.001 & 10.816$\pm$0.020 & \textbf{100.00}* & 97.62* & 98.10* \\
        & GraphDF+FC & 93.88$\pm$4.76 & & 0.064$\pm$0.000 & 10.928$\pm$0.038 & \textbf{100.00}$\pm$0.00 & 98.58$\pm$0.25 & \textbf{98.54}$\pm$0.48 \\
    \midrule
        \multirow{5.5}{*}{One-shot}
        & MoFlow~\cite{zang2020moflow} & 91.36$\pm$1.23 & & 0.017$\pm$0.003 & 4.467$\pm$0.595 & \textbf{100.00}$\pm$0.00 & 98.65$\pm$0.57 & 94.72$\pm$0.77 \\
        & EDP-GNN~\cite{score-based/graph/1} & 47.52$\pm$3.60 & & 0.005$\pm$0.001 & \textbf{2.680}$\pm$0.221 & \textbf{100.00}$\pm$0.00 & \textbf{99.25}$\pm$0.05 & 86.58$\pm$1.85 \\
        & GraphEBM~\cite{liu2021graphebm} & 8.22$\pm$2.24 & & 0.030$\pm$0.004 & 6.143$\pm$0.411 & \textbf{100.00}$\pm$0.00* & 97.90$\pm$0.14* & 97.01$\pm$0.17* \\
    \cmidrule(l{2pt}r{2pt}){2-9}
        & \textbf{GDSS-seq} (Ours) & 94.47$\pm$1.03 & & 0.010$\pm$0.001 & 4.004$\pm$0.166 & \textbf{100.00}$\pm$0.00 & 94.62$\pm$1.40 & 85.48$\pm$1.01 \\
        & \textbf{GDSS} (Ours) & \textbf{95.72}$\pm$1.94 & & \textbf{0.003}$\pm$0.000 & 2.900$\pm$0.282 & \textbf{100.00}$\pm$0.00 & 98.46$\pm$0.61 & 86.27$\pm$2.29 \\
    \bottomrule
    \end{tabular}}
    \label{tab:qm9}
\vspace{0.05in}
\end{table*}

%% file: table/zinc250k.tex
\begin{table*}[t!]
    \caption{\small \textbf{Generation results on the ZINC250k dataset.} Results are the means and the standard deviations of 3 runs. Values denoted by * are taken from the respective original papers. Other results are obtained by running open-source codes. Best results are marked as bold.}
\vspace{-0.1in}
    \centering
    \resizebox{\textwidth}{!}{
    \begin{tabular}{llc@{\hskip 0.01in}cccccc}
    \toprule
        & \multirow{2}{*}{Method} & Validity w/o & \multirow{2}{*}{$\uparrow$} & NSPDK \multirow{2}{*}{$\downarrow$} & \multirow{2}{*}{FCD $\downarrow$} & \multirow{2}{*}{Validity (\%) $\uparrow$} & \multirow{2}{*}{Uniqueness (\%) $\uparrow$} & \multirow{2}{*}{Novelty (\%) $\uparrow$} \\
        & & correction (\%) & & MMD & & & & \\
    \midrule
        \multirow{4}{*}{Autoreg.}
        & GraphAF~\cite{shi2020graphaf} & 68* & & 0.044$\pm$0.006 & 16.289$\pm$0.482 & \textbf{100.00}* & 99.10* & \textbf{100.00}* \\
        & GraphAF+FC & 68.47$\pm$0.99 & & 0.044$\pm$0.005 & 16.023$\pm$0.451 & \textbf{100.00}$\pm$0.00 & 98.64$\pm$0.69 & 99.99$\pm$0.01 \\
        & GraphDF~\cite{luo2021graphdf} & 89.03* & & 0.176$\pm$0.001 & 34.202$\pm$0.160 & \textbf{100.00}* & 99.16* & \textbf{100.00}* \\
        & GraphDF+FC & 90.61$\pm$4.30 & & 0.177$\pm$0.001 & 33.546$\pm$0.150 & \textbf{100.00}$\pm$0.00 & 99.63$\pm$0.01 & \textbf{100.00}$\pm$0.00 \\
    \midrule
        \multirow{5.5}{*}{One-shot}
        & MoFlow~\cite{zang2020moflow} & 63.11$\pm$5.17 & & 0.046$\pm$0.002 & 20.931$\pm$0.184 & \textbf{100.00}$\pm$0.00 & \textbf{99.99}$\pm$0.01 & \textbf{100.00}$\pm$0.00 \\
        & EDP-GNN~\cite{score-based/graph/1} & 82.97$\pm$2.73 & & 0.049$\pm$0.006 & 16.737$\pm$1.300 & \textbf{100.00}$\pm$0.00 & 99.79$\pm$0.08 & \textbf{100.00}$\pm$0.00 \\
        & GraphEBM~\cite{liu2021graphebm} & 5.29$\pm$3.83 & & 0.212$\pm$0.075 & 35.471$\pm$5.331 & 99.96$\pm$0.02* & 98.79$\pm$0.15* & \textbf{100.00}$\pm$0.00* \\
    \cmidrule(l{2pt}r{2pt}){2-9}
        & \textbf{GDSS-seq} (Ours) & 92.39$\pm$2.72 & & 0.030$\pm$0.003 & 16.847$\pm$0.097 & \textbf{100.00}$\pm$0.00 & 99.94$\pm$0.02 & \textbf{100.00}$\pm$0.00 \\
        & \textbf{GDSS} (Ours) & \textbf{97.01}$\pm$0.77 & & \textbf{0.019}$\pm$0.001 & \textbf{14.656}$\pm$0.680 & \textbf{100.00}$\pm$0.00 & 99.64$\pm$0.13 & \textbf{100.00}$\pm$0.00 \\
    \bottomrule
    \end{tabular}}
    \label{tab:zinc250k}
\vspace{0.05in}
\end{table*}

%% file: table/solver_appendix.tex
\begin{table*}[t!]
    \caption{\small \textbf{Comparison between fixed step size SDE solvers.} We additionally provide the results of the proposed S4 solver on other datasets not included in the table of Figure~\ref{fig:vis}. Best results are highlighted in bold (smaller the better).}
    \vspace{-0.1in}
    \centering
    \resizebox{\textwidth}{!}{
    \renewcommand{\arraystretch}{1.0}
    \renewcommand{\tabcolsep}{6pt}
    \begin{tabular}{l l cccc cccc cccc cccc}
    \toprule
        & \multicolumn{4}{c}{{Ego-small}} & \multicolumn{4}{c}{{Grid}} & \multicolumn{4}{c}{{QM9}} & \multicolumn{4}{c}{{ZINC250k}}\\
    \cmidrule(l{2pt}r{2pt}){2-5}
    \cmidrule(l{2pt}r{2pt}){6-9}
    \cmidrule(l{2pt}r{2pt}){10-13}
    \cmidrule(l{2pt}r{2pt}){14-17}
        Solver & Deg. & Clus. & Orbit & Time (s) & Deg. & Clus. & Orbit & Time (s) & Val. w/o corr. (\%) & NSPDK & FCD & Time (s) & Val. w/o corr. (\%) & NSPDK & FCD & Time (s) \\
    \midrule
        EM & \textbf{0.021} & \textbf{0.024} & \textbf{0.007} & \textbf{40.31} & 0.278 & 0.008 & 0.089 & \textbf{235.36} & 67.44 & 0.016 & 4.809 & 63.87 & 15.92 & 0.086 & 26.049 & \textbf{1019.89} \\
        Reverse & 0.032 & 0.046 & 0.010 & 41.55 & 0.278 & 0.008 & 0.089 & 249.86 & 69.32 & 0.016 & 4.823 & 65.23 & 46.02 & 0.052 & 21.486 & 1021.09 \\
        \midrule
        EM + Langevin & 0.032 & 0.040 & 0.009 & 78.17 & \textbf{0.111} & \textbf{0.005} & \textbf{0.070} & 483.47 & 93.98 & 0.008 & 3.889 & 111.49 & 94.47 & 0.025 & 15.292 & 2020.87 \\
        Rev. + Langevin & 0.032 & 0.046 & 0.021 & 77.82 & \textbf{0.111} & \textbf{0.005} & \textbf{0.070} & 500.01 & \textbf{95.72} & \textbf{0.003} & 2.900 & 114.57 & \textbf{97.01} & \textbf{0.019} & 14.656 & 2020.06 \\
        \midrule
        \textbf{S4} (Ours) & 0.032 & 0.044 & 0.009 & 41.25 & 0.125 & 0.008 & 0.076 & 256.24 & 95.13 & \textbf{0.003} & \textbf{2.777} & \textbf{63.78} & 95.52 & 0.021 & \textbf{14.537} & 1021.21 \\
    \bottomrule
    \end{tabular}}
    \label{tab:S4_full}
\vspace{-0.1in}
\end{table*}